\documentclass[11pt]{article}

\usepackage[final]{acl}

\usepackage{times}
\usepackage{latexsym}
\usepackage{booktabs}
\usepackage{multirow}
\usepackage{graphicx}
\usepackage[T1]{fontenc}
\usepackage[most]{tcolorbox}
\tcbuselibrary{breakable, skins}
\usepackage{xcolor}
\usepackage[table]{xcolor}
\definecolor{myBack}{HTML}{FEF8FA}
\definecolor{myFrame}{HTML}{D2263E}

\newcommand{\hl}[1]{\cellcolor{orange!50}\textbf{#1}}
\usepackage[utf8]{inputenc}
\usepackage{pifont}
\usepackage{amsmath,amssymb,amsthm}
\usepackage{adjustbox}
\usepackage{subcaption}
\usepackage{cleveref}
\usepackage{microtype}
\usepackage{hyperref}
\usepackage{inconsolata}
\usepackage{threeparttable}
\usepackage{booktabs}
\usepackage{array}
\usepackage{float}
\usepackage{amsmath}
\usepackage{amssymb}
\usepackage{amsthm}
\newtheorem{theorem}{Theorem}
\usepackage{bm}
\usepackage{xurl}
\usepackage[linesnumbered,ruled,vlined]{algorithm2e}

\title{CIA: Inferring the Communication Topology from LLM-based Multi-Agent Systems}

\author{
Yongxuan Wu$^{1,2}$, Xixun Lin$^{1,2}$\thanks{Corresponding author.}, He Zhang$^{3}$, Nan Sun$^{1,2}$, Kun Wang$^{4}$, \\ \textbf{Chuan Zhou$^{5}$}, \textbf{Shirui Pan$^{3}$}, \textbf{Yanan Cao$^{1,2}$}\\
 $^1$ Institute of Information Engineering, Chinese Academy of Sciences, Beijing, China \\
 $^2$ School of Cyber Security, University of Chinese Academy of Sciences, Beijing, China \\
 $^3$ Griffith University, Brisbane, Australia\\
 $^4$ Nanyang Technological University, Singapore\\
 $^5$ Academy of Mathematics and Systems Science, Chinese Academy of Sciences, Beijing, China\\
\texttt{\{wuyongxuan,linxixun\}@iie.ac.cn}\\
}

\begin{document}
\maketitle
\begin{abstract}
LLM-based Multi-Agent Systems (MAS) have demonstrated remarkable capabilities in solving complex tasks. Central to MAS is the communication topology which governs how agents exchange information internally. Consequently, the security of communication topologies has attracted increasing attention. In this paper, we investigate a critical privacy risk: MAS communication topologies can be inferred under a restrictive black-box setting, exposing system vulnerabilities and posing significant intellectual property threats. To explore this risk, we propose \textbf{Communication Inference Attack (CIA)}, a novel attack that constructs new adversarial queries to induce intermediate agents’ reasoning outputs and models their semantic correlations through the proposed global bias disentanglement and LLM-guided weak supervision. Extensive experiments on MAS with optimized communication topologies demonstrate the effectiveness of CIA, achieving an average AUC of 0.87 and a peak AUC of up to 0.99, thereby revealing the substantial privacy risk in MAS. The source code is available at \url{https://github.com/aabbbcd/CIA}.
\end{abstract}

\section{Introduction}
LLM-based agents have rapidly evolved into powerful intelligent systems, exhibiting human-like capabilities in cognition and reasoning~\cite{shinn2023reflexion,jin2023surrealdriver,yang2024swe}. To further scale these capabilities, recent research has shifted toward LLM-based multi-agent systems (MAS). By orchestrating the collaboration among multiple agents, MAS can tackle complex tasks that are beyond the reach of a single agent~\cite{wang2024unleashing,zhangmore,wang2025megaagent}. As a result, MAS have demonstrated remarkable performance across various domains, including software engineering~\cite{he2025llm}, scientific discovery~\cite{ghafarollahi2025sciagents}, and social simulation~\cite{taillandier2025integrating}.
\par
This advantage primarily stems from the optimized communication topology within MAS, which enables agents to exchange information and refine their decisions through collaboration or debate. Along with the rapid advancement of MAS, their security has attracted increasing attention~\cite{wang2024badagent,li2025commercial,yan2025attack,DBLP:journals/corr/abs-2509-18970}. In terms of adversarial attacks, existing studies mainly focus on inducing toxic outputs or misinformation spread among agents through various attack strategies~\cite{lee2024prompt,yu2025infecting,he2025red}. In this paper, we investigate a largely underexplored yet critical privacy risk: \textbf{Is the communication topology of MAS itself vulnerable to leakage?}
\begin{figure}[t]
        \centering
        \includegraphics[width=0.48\textwidth]{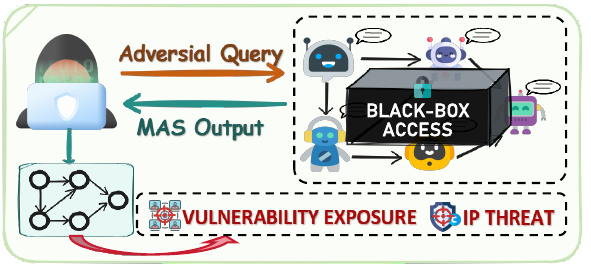}
        \caption{\textbf{Illustration of inferring the communication topology.} The adversary infers the communication topology under a black-box setting, resulting in a severe \textit{Vulnerability Exposure} that enables adversaries to pinpoint critical agents for targeted attacks and \textit{IP Threat} that leaks valuable proprietary assets of developers.}
        \label{fig:sub1}
\end{figure} 
\par
Compared with previous attacks, this is a stealthier privacy risk, as the adversary does not aim to disrupt task execution but seeks to infer the internal information of MAS solely through black-box queries. Moreover, as illustrated in~\Cref{fig:sub1}, once the communication topology is inferred, the consequences can be severe. 
First, it leads to \textit{Vulnerability Exposure}. Revealing the communication topology of MAS exposes the systems’ internal organization, allowing the adversary to identify critical or weak agents for targeted jailbreaking or instruction injection, thereby compromising MAS at low cost~\cite{raza2025trism}. 
Second, it poses a significant \textit{IP Threat}. A highly optimized communication topology encapsulates substantial computational resources and expert knowledge, representing valuable proprietary assets~\cite{zhang2024g,li2025adaptive}. The leakage of this topology constitutes a direct infringement of IP, consequently undermining the developers' competitive advantage~\cite{kong2025survey}. 

\par
To explore this privacy risk, we propose \textbf{Communication Inference Attack (CIA)}, a novel attack for inferring the communication topology within MAS. CIA operates under a practical black-box setting, where the adversary is neither authorized to access nor alter any internal information of MAS. Instead, the adversary merely interacts with MAS by issuing queries and observing their final responses. CIA consists of two stages: \textit{Reasoning Output Induction} and \textit{Semantic Correlations Modeling}. In the first stage, CIA crafts adversarial queries to induce the final output to reveal the intermediate agents' reasoning outputs. In the second stage, we introduce global bias disentanglement and LLM-guided weak supervision to mitigate spurious correlations and enhance the topological information embedded in these reasoning outputs, and subsequently analyze their semantic correlations to infer the communication topology. Extensive evaluations demonstrate the effectiveness of CIA, revealing the substantial privacy risk in the communication topology of MAS. In summary, our contributions are as follows.

\begin{itemize}
    \item We investigate a largely underexplored privacy risk in MAS: the vulnerability of their communication topologies to being inferred under a black-box setting, which poses significant IP threats and vulnerability exposure.  
    \item We propose the CIA, a novel attack that first crafts adversarial queries to expose the reasoning outputs of intermediate agents and then models the semantic correlations of these outputs using global bias disentanglement and LLM-guided weak supervision to infer the confidential communication topology.
    \item We conduct experiments on MAS built using well-optimized communication topologies across multiple task scenarios. 
    Experimental results show that the communication topology in MAS can be effectively inferred, with CIA achieving an average AUC of 0.87 and a peak AUC of up to 0.99.
\end{itemize}

\section{Related Work}
\label{Related_Works}
\paragraph{Topology Design for MAS.} 
The communication topology is fundamental for the effectiveness of MAS, serving as the backbone of collective intelligence and joint reasoning~\cite{cemri2025multi}. Consequently, much work has focused on designing communication topologies for MAS~\cite{liu2025graph}. Early designs rely on handcrafted or heuristic patterns that lack the flexibility to adapt to diverse tasks~\cite{hong2023metagpt,li2023camel,qian2024chatdev}. To overcome this limitation, recent methods have introduced \emph{Generative Optimization Strategies} to dynamically generate agent compositions or communication topologies tailored to specific tasks~\cite{zhang2024g, zhang2024aflow, li2025adaptive}. These approaches not only achieve state-of-the-art (SOTA) performance but also reduce the resource costs of redundant communications in MAS.

\paragraph{Adversarial Attacks against MAS.}
 Recent research on adversarial attacks against MAS has primarily focused on inducing toxic outputs or spreading misinformation~\cite{yu2025survey}. Specifically, some methods study communication content-based attacks, such as task abandonment~\cite{amayuelas2024multiagent}, communication tampering~\cite{he2025red}, or malicious prompt propagation~\cite{lee2024prompt,yu2025infecting}. Meanwhile, some approaches explore communication topology-based attacks by evaluating the resilience of different communication topologies to identify which topologies are more vulnerable to adversarial attacks~\cite{huang2024resilience, yu2025netsafe}. However, the privacy risk of the communication topology itself remains largely unexplored. In this paper, we focus on this important risk and investigate whether the communication topology of MAS can be inferred in a black-box setting. 
inferring edges by exploiting correlations in prediction posteriors or gradient information.
\begin{figure*}[htbp]
    \centering
    \includegraphics[width=\linewidth]{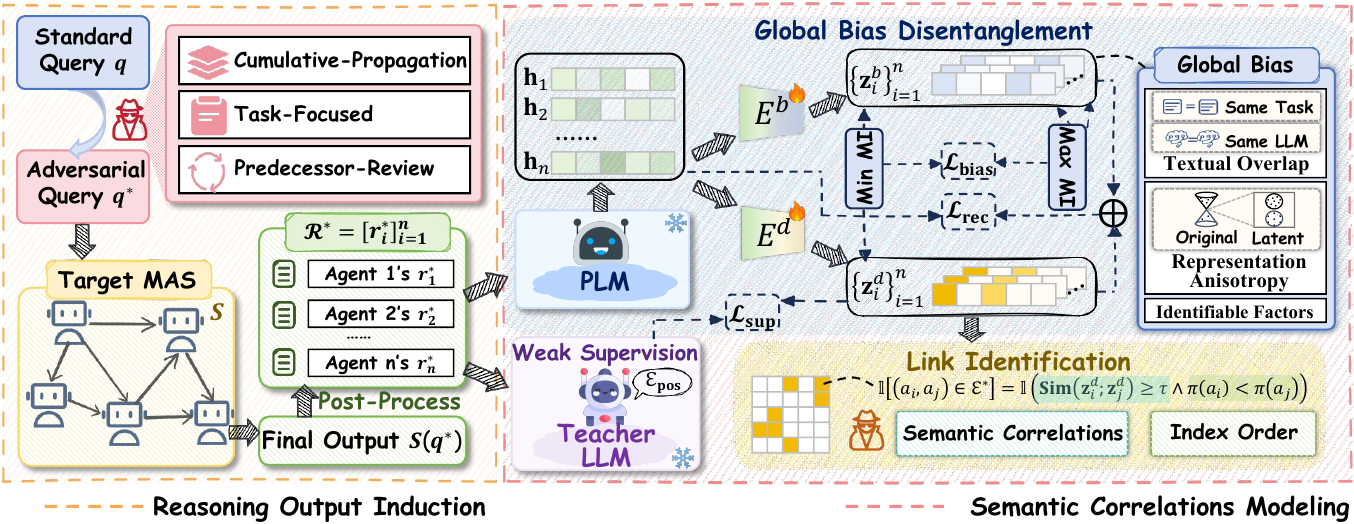} 
    \caption{The overview of CIA. CIA first induces the final output that reveals the reasoning outputs of intermediate agents by crafting adversarial queries. CIA then infers the communication topology by modeling semantic correlations among these outputs using global bias disentanglement and LLM-guided weak supervision.} 
    \label{fig:framework}
\end{figure*}
\section{Background}
We introduce a basic framework of LLM-based MAS. Let $\mathcal{S}=(\mathcal{P},\mathcal{G})$ represent MAS. Here, $\mathcal{P}=\{p_i\}_{i=1}^n$ denotes the set of agent profiles, each of which includes the agent's system prompt, callable tools, and other configuration details; $\mathcal{G} = (\mathcal{A}, \mathcal{E})$ denotes the communication topology of $\mathcal{S}$, which is modeled as a directed acyclic graph (DAG) capturing the information flow for task completion~\cite{zhang2024g,li2025assemble}. $\mathcal{A}=\{a_i\}_{i=1}^n$ denotes the set of agents, each corresponding to an LLM, and $\mathcal{E} \subseteq \mathcal{A} \times \mathcal{A}$ denotes the set of directed edges. A directed edge $e_{j \rightarrow i} = (a_j, a_i) \in \mathcal{E}$ indicates that agent $a_i$ is a designated recipient of information from agent $a_j$.

Based on this framework, for the given task with the corresponding query $q$, the $i$-th agent $a_i$ generates its reasoning output $r_i$ as follows: 
\begin{equation}
r_i = \mathrm{LLM}(p_i, q, \mathcal{O}_i).
\end{equation}
$\mathcal{O}_i$ is the set of outputs generated by the predecessor agents of $a_i$, which can be defined as
\begin{equation}
\mathcal{O}_i = \{\, r_j \mid a_j \in \mathcal{N}_{\mathrm{in}}(i) \,\},
\end{equation}
where $\mathcal{N}_{\mathrm{in}}(i) = \{\, a_j \mid (a_j, a_i) \in \mathcal{E} \,\}$
denotes the set of predecessor agents of $a_i$. The final output of $\mathcal{S}$ is produced by the decision agent\footnote{This formulation follows a common abstraction in MAS, where the final output is generated by a decision agent via summarization or majority voting.}:
\begin{equation}
r_n = \mathcal{S}(q) = \mathrm{LLM}(p_n, q, \mathcal{O}_n).
\end{equation}
From the above formulation, we obtain that the effective communication, governed explicitly by $\mathcal{G}$, is pivotal to the performance of $\mathcal{S}$, as it facilitates the efficient exchange and propagation of information among agents for task completion.

\section{Research Problem}
To investigate the privacy risk of communication topology leakage, we propose the Communication Inference Attack (CIA). Under our attack scenario, we introduce the system model, the adversary goal, and the adversary capabilities as follows.

\paragraph{System Model.} We consider the MAS $\mathcal{S}$ designed to handle complex tasks, such as mathematical reasoning and code generation. In a standard usage scenario, a user provides a query $q$ to the system, and $\mathcal{S}$ returns an output $\mathcal{S}(q)$ generated through collaborative interactions among agents.

\paragraph{Attack Goal.} The adversary aims to infer the communication topology ${\mathcal{G}}$ through querying $\mathcal{S}$ and analyzing final outputs. This attack exposes MAS vulnerability by revealing its internal organization, enabling more targeted attacks on critical agents and threatens developers’ IP. 

\paragraph{Adversary Capabilities.} The adversary operates under a practical black-box setting. This implies that the adversary can only interact with $\mathcal{S}$ through its external interface. The adversary has no access to the internal information of ${\mathcal{S}}$, such as reasoning traces, agent profiles, and system configurations.

\section{Methodology}
The intuition underlying CIA is that agents in $\mathcal{S}$ do not operate independently; instead, each agent's output is conditioned on the responses of its predecessors, resulting in stronger semantic dependencies between agents with direct topological connections than between those without such connections. Under the black-box setting, CIA can only observe the final output $r_n$, as the internal information is inaccessible. Consequently, CIA first aims to induce $\mathcal{S}$ to reveal the reasoning outputs of intermediate agents. Here, intermediate agents refer to the agents in $\mathcal{S}$ excluding the decision agent. They participate in the intermediate reasoning process, but their outputs are not available. CIA then analyzes the semantic correlations between these outputs to infer the communication topology within $\mathcal{S}$. 
\par
Following this intuition, CIA naturally owns two stages. \textbf{1) Reasoning Output Induction}: CIA constructs  adversarial queries to interact with the target MAS $\mathcal{S}$, inducing the final output that reveals reasoning outputs of intermediate agents. \textbf{2) Semantic Correlations Modeling}: CIA infers $\mathcal{G}$ by modeling the semantic correlations among these outputs. The overview of CIA is shown in \Cref{fig:framework}.

\subsection{Reasoning Output Induction}\label{outputs}
This section presents a novel adversarial querying strategy for eliciting intermediate agents’ reasoning from the final output.
Concretely, the adversarial query imposes three specific constraints on each agent’s output. 

\ding{182} \textbf{Cumulative-Propagation Constraint.}
To ensure the final output contains the reasoning outputs of intermediate agents, we impose the cumulative-propagation constraint, requiring each agent to copy the historical record of its predecessors and append their reasoning outputs as the updated history. Through this cumulative recording process, the reasoning outputs are propagated through $\mathcal{G}$. The template for this constraint is as follows:

\begin{tcolorbox}[
    colback=myBack,
    colframe=myFrame,
    boxrule=0.8pt,
    width=\linewidth,      
    breakable,             
    arc=2pt,
    left=2pt, right=2pt,
    title=\raisebox{-0.3em}{\includegraphics[height=1.1em]{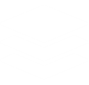}} Cumulative-Propagation Constraint.
]
1.\textbf{Goal}: Carry the history reasoning output.

2.\textbf{Action}: Copy the content of the previous agents’ [PREVIOUS HISTORY] (if any) and append the previous agents’ [REASONING OUTPUT] content.

3.\textbf{Format}: <Previous Agent [PREVIOUS HISTORY] Content> \textbf{|||} <Previous Agent [REASONING OUTPUT] Content>.
\end{tcolorbox}

\ding{183} \textbf{Task-Focused Constraint.} 
The adversarial query inevitably introduces task-irrelevant information that can distract agents and cause deviations from their original reasoning trajectories. To mitigate this effect, we impose the task-focused constraint 
that requires each agent to focus exclusively on the task-relevant fields explicitly marked in the input and the reasoning outputs of its predecessors. The template for this constraint is as follows:

\begin{tcolorbox}[
    colback=myBack,
    colframe=myFrame,
    boxrule=0.8pt,
    width=\linewidth,      
    breakable,             
    arc=2pt,
    left=2pt, right=2pt,
    title=\raisebox{-0.4em}{\includegraphics[height=1.4em]{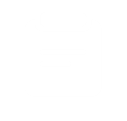}} Task-Focused Constraint.
]
1.\textbf{Goal}: Prevent reasoning deviation caused by task-irrelevant information.

2.\textbf{Action}: Exclusively extract and analyze the [TASK] and the previous agents' [REASONING OUTPUT] during your reasoning, disregarding any other information provided in the input.
\end{tcolorbox}
\ding{184} \textbf{Predecessor-Review Constraint.} 
To further strengthen the semantic correlations between the reasoning outputs of adjacent agents, we impose the predecessor-review constraint on each agent to review the predecessor agents’ reasoning outputs before generating its own output. The template for this constraint is as follows:
\begin{tcolorbox}[
    colback=myBack,
    colframe=myFrame,
    boxrule=0.8pt,
    width=\linewidth,      
    breakable,             
    arc=2pt,
    left=2pt, right=2pt,
    title=\raisebox{-0.4em}{\includegraphics[height=1.4em]{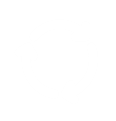}} Predecessor-Review Constraint.
]
1.\textbf{Goal}: Strengthen the semantic association between predecessor and successor agents.

2.\textbf{Action}: Prior to your reasoning, explicitly review the previous agents' [REASONING OUTPUT] and incorporate them in your own [REASONING OUTPUT].
\end{tcolorbox}
Guided by these three constraints, we use the adversarial query $q^{*}$ to interact with $\mathcal{S}$ for generating $\mathcal{S}(q^{*})$ that reveals the reasoning outputs of intermediate agents. Since $\mathcal{S}(q^*)$ is an unstructured text, we need to separate the outputs of each agent for the downstream semantic correlations modeling. Therefore, we post-process $\mathcal{S}(q^*)$ into a list, denoted as $\mathcal{R}^* = [r_i^*]_{i=1}^n$, where $r_i^*$ corresponds to the reasoning output of $i$-th agent. This list $\mathcal{R}^*$ follows the order in which agents complete their reasoning task, implicitly reflecting the communication direction. Detailed procedures for this post-processing are provided in Appendix~\ref{app:post}.

\subsection{Semantic Correlations Modeling}
With the recovered agent outputs, we aim to infer $\mathcal{G}$ by modeling their semantic correlations in the following steps. First, we propose the global bias disentanglement to learn debiased representations for removing the spurious information in $\mathcal{R}^*$. Second, we design an LLM-guided weak supervision to refine these debiased representations, enhancing their capability to learn the topological information of $\mathcal{G}$. Finally, we identify whether a link exists between agents by computing the similarity between their refined representations. 

\paragraph{Global Bias Disentanglement (GBD).}\label{para:gbd}
In fact, agents may still exhibit strong semantic similarity in their reasoning outputs even without explicit communication, rendering the semantic correlations among these recovered outputs can be highly spurious. Such spurious correlations stem from multiple sources. For instance, agents may share the same base LLM and operate on the same task, which naturally results in overlapping content and similar linguistic patterns~\cite{bommasani2021opportunities}. In addition, due to representation anisotropy~\cite{godey2024anisotropy}, agents may produce semantically distinct outputs that nonetheless appear highly correlated in the embedding space. Beyond these identifiable factors, other unobservable sources may further exacerbate this issue~\cite{KishoreChakrabarty2025CausalII}.
\par
We collectively refer to the sources that induce spurious correlations as \textbf{Global Bias}, since they represent the spurious information that is globally shared across agents’ reasoning outputs. Such global bias can mislead the adversary into focusing on semantic signals that are unrelated to the communication topology, thereby inflating pairwise similarities among agents. As a consequence, many non-communicating agent pairs are falsely inferred as having communicated. 
\par
To mitigate the impact of global bias, we propose GBD to learn debiased representations. Specifically, we first employ a pretrained language model\footnote{We utilize the \texttt{all-MiniLM-L6-v2} in implementation.} $f_\theta$ to encode $\mathcal{R}^*$. For the $i$-th agent with its reasoning output $r_i^*$, $f_\theta$ produces an initial representation $\mathbf{h}_i$. We then project each $\mathbf{h}_i$ into two latent subspaces via two trainable encoders, $E^d$ and $E^b$, representing the debiased encoder and the bias encoder, respectively:
\begin{equation}
\mathbf{z}_i^d = E^d(\mathbf{h}_i), \quad \mathbf{z}_i^b = E^b(\mathbf{h}_i),
\label{zz}
\end{equation}
where $\mathbf{z}_i^d$ and $\mathbf{z}_i^b$ denote the debiased and biased representations for $r_i^*$, respectively. 
\par
Since the global bias is shared across these reasoning outputs, we can maximize the mutual information among $\{\mathbf{z}_i^b\}_{i=1}^{n}$ to encourage $E^b$ to effectively capture the global bias. Meanwhile, to prevent such global bias from influencing $\{\mathbf{z}_i^d\}_{i=1}^{n}$, we simultaneously minimize the mutual information between $\mathbf{z}_i^d$ and $\mathbf{z}_i^b$ for each agent. These two information-theoretic objectives are jointly optimized via the following loss:
\begin{equation}
\mathcal{L}_{\mathrm{bias}}=- \ \mathcal{I}(\mathbf{z}_1^b; \dots; \mathbf{z}_n^b)+\sum_{i=1}^n\ \mathcal{I}(\mathbf{z}_i^d; \mathbf{z}_i^b).
\label{loss_bias}
\end{equation}
The computational details of Eq.(\ref{loss_bias}) is given in Appendix~\ref{app:TC}. 
\par
To prevent the encoded information from being lost during disentanglement~\cite{bousmalis2016domain}, we also introduce a reconstruction loss in GBD. Specifically, for each $\mathbf{h}_i$ with its disentangled $\mathbf{z}_i^d$ and $\mathbf{z}_i^b$, we concatenate them together and feed it into a decoder $D$ to reconstruct $\mathbf{h}_i$: 
\begin{equation}
\hat{\mathbf{h}}_i = D(\mathbf{z}_i^d \oplus \mathbf{z}_i^b).
\end{equation}
The reconstruction loss is defined as
\begin{equation}
\mathcal{L}_{\mathrm{rec}}
=
\sum_{i=1}^{n}
\left\|
\mathbf{h}_i - \hat{\mathbf{h}}_i
\right\|_2^2.
\label{rec_loss}
\end{equation}
Finally, the overall training loss for GBD is

\begin{equation}
\begin{aligned}
\mathcal{L}_{\mathrm{GBD}}
= \mathcal{L}_{\mathrm{rec}}
+\mathcal{L}_{\mathrm{bias}}.
\end{aligned}
\label{dis}
\end{equation}
\paragraph{LLM-guided Weak Supervision (LWS).}
\label{para:ws}
At the above step, $\mathbf{z}_i^d$ is learned only from the textual information of $r_i^*$. We aim to enable $\mathbf{z}_i^d$ to capture the information at the structural level of $\mathcal{G}$, facilitating the more accurate communication inference. However, such information is not directly accessible. To this end, we leverage the information inferred by a teacher LLM\footnote{We utilize GPT-5 in our implementation.} as weak supervision signals, distilling the structural knowledge into $\mathbf{z}_i^d$.
\par
Given $\mathcal{R^*}$, the teacher LLM is prompted to infer the top-$k$ communication edges with the highest confidence scores (The exact prompt template is provided in Appendix~\ref{app:prompts Weak Supervision}.). We denote the edges inferred by this LLM as the positive set $\mathcal{E}_{\mathrm{pos}}$, and sample negative pairs $\mathcal{E}_{\mathrm{neg}}$ from the remaining agent pairs outside $\mathcal{E}_{\mathrm{pos}}$. As the LLM-inferred edges may be noisy and the remaining pairs are not guaranteed to be true negatives, we adopt the trick of label smoothing~\cite{dettmers2018convolutional} and define the weak supervision loss as
\begin{equation} 
\begin{aligned} 
\mathcal{L}_{\mathrm{LWS}} &= - \frac{1}{|\mathcal{E}_{\mathrm{pos}}|} \sum_{(a_i,a_j) \in \mathcal{E}_{\mathrm{pos}}} \mathcal{L}_{\mathrm{pos}}(a_i,a_j) \\ &\quad - \frac{1}{|\mathcal{E}_{\mathrm{neg}}|} \sum_{(a_u,a_v) \in \mathcal{E}_{\mathrm{neg}}} \mathcal{L}_{\mathrm{neg}}(a_u,a_v), 
\end{aligned} 
\label{sup}
\end{equation}
where $\mathcal{L}_{\mathrm{pos}}$ and $\mathcal{L}_{\mathrm{neg}}$ are the label-smoothed binary cross-entropy losses computed based on the similarity between the debiased representations for each positive and negative agent pair, respectively. Detailed formulations of $\mathcal{L}_{\mathrm{pos}}$ and $\mathcal{L}_{\mathrm{neg}}$ are provided in Appendix~\ref{pos}.
\par
The trainable modules in CIA are $E^d$ and $E^b$, and the final objective of CIA is to minimize the total loss:

\begin{equation}
\begin{aligned}
\mathcal{L}_{\mathrm{CIA}}=\mathcal{L}_{\mathrm{GBD}}+\mathcal{L}_{\mathrm{LWS}}.
\end{aligned}
\end{equation}

\paragraph{Link Identification.} After training, the communication topology $\mathcal{G}$ can be identified from the debiased representations. The existence of an edge between agents $a_i$ and $a_j$ is determined by the similarity between $\mathbf{z}_i^d$ and $\mathbf{z}_j^d$, while the edge direction is inferred according to their relative order in $\mathcal{R}^*$:

\begin{equation}
\begin{aligned}
\mathbb{I}[(a_i, a_j) \in \mathcal{E}]
&= \mathbb{I}\!\Big(
    \mathrm{Sim}(\mathbf{z}_i^d, \mathbf{z}_j^d) \ge \tau 
    \;\land\; \\
&\qquad
    \pi(a_i) < \pi(a_j)
\Big),
\end{aligned}
\label{eq:link}
\end{equation}
where $\mathrm{Sim}(\cdot,\cdot)$ indicates a distance-based similarity function, $\tau$ is a threshold, and $\pi(a)$ denote the index of agent $a$’s reasoning output in $\mathcal{R^*}$.

\begin{table*}[t]
  \centering
  \resizebox{\textwidth}{!}{ 
    \begin{tabular}{llcccccccccccc}
    \toprule
    \multirow{2}{*}{\textbf{Dataset}} & \multirow{2}{*}{\textbf{Method}} & \multicolumn{3}{c}{\textbf{MMLU}} & \multicolumn{3}{c}{\textbf{GSM8K}} & \multicolumn{3}{c}{\textbf{SVAMP}} & \multicolumn{3}{c}{\textbf{HumanEval}} \\
    \cmidrule(lr){3-5} \cmidrule(lr){6-8} \cmidrule(lr){9-11} \cmidrule(lr){12-14}
          &       & AUC   & ACC   & F1    & AUC   & ACC   & F1    & AUC   & ACC   & F1 & AUC   & ACC   & F1\\
    \midrule
    \multirow{6}{*}{\textbf{G-Designer}} 
          & GPT-5 &  0.5833     &  0.8033     &0.3084     &  0.6274    & 0.5887   &  0.4555    &0.6272      &0.7876       &0.6815        & 0.6201 &0.5211       & 0.3204     \\ \cmidrule{2-14}
          & Gemini-2.5-Pro &   0.6869     &  0.7900     & 0.3697          & 0.6152      &   0.5728    & 0.4545   &  0.6300     &0.7600    & 0.7059   & 0.6318      &  0.5541     &0.3899\\\cmidrule{2-14}
          & Llama-3.1-8B-Instruct & 0.5995      & 0.7157      &0.4156       & 0.5284      & 0.5348      &0.4689       & 0.4567  &0.6252       & 0.4361  &0.4922       &  0.4626     &0.2863\\\cmidrule{2-14}
          & Mistral-7B-Instruct-v0.2 & 0.5192      & 0.7555      &0.2889       &0.4105       &0.4260       &0.2866       &0.4977       & 0.6704      &0.4351  &0.4893       &0.4430       &0.2811\\\cmidrule{2-14}

          & \textbf{CIA (Ours)} &  \hl{0.8324}     &    \hl{0.8147}   &   \hl{0.7761}    & \hl{0.8585}      & \hl{0.8382}      & \hl{0.7744}      &\hl{0.8561}  &\hl{0.8083}       &\hl{0.7682}        &\hl{0.7594}       &\hl{0.7362}       &\hl{0.6566}\\
    \midrule
    \multirow{6}{*}{\textbf{AGP}} 
          & GPT-5 & 0.6577      &  0.6974     &  0.5472     &  0.6233     & 0.5972      & 0.5357      &  0.6199     & 0.7940       & 0.6330 & 0.4433      &0.6661       & 0.2689\\\cmidrule{2-14}
          & Gemini-2.5-Pro & 0.6324      &   0.5935    &  0.5217     &  0.6177     & 0.5733      & 0.5174      &   0.6087  & 0.7733     & 0.6241 & 0.4033      &  0.6122     &0.2158 \\\cmidrule{2-14}
          & Llama-3.1-8B-Instruct & 0.4812      & 0.6581      &0.4786       & 0.4265      &0.5190       & 0.5100&  0.4355     &0.6315       & 0.5396      &0.3961  & 0.6450            &0.2500 \\\cmidrule{2-14}
          & Mistral-7B-Instruct-v0.2 & 0.5076      & 0.6579      &0.4246       &0.4867       &0.6500       & 0.2866      &  0.4939     & 0.6588      &0.4475  & 0.5000      & 0.6722      & 0.2857\\\cmidrule{2-14}
        
          & \textbf{CIA (Ours)} & \hl{0.8411}      &\hl{0.7942}       &\hl{0.7567}       &\hl{0.8804}       &\hl{0.8204}       &  \hl{0.7866}     & \hl{0.8979}       & \hl{0.8380}       &  \hl{0.7923} &   \hl{0.8100}   & \hl{0.7777}     &\hl{0.6759} \\
    \midrule
    \multirow{6}{*}{\textbf{ARG-Designer}} 
          & GPT-5 & 0.5879      &  0.7545& 0.4094      &  0.6984     &0.7672       &0.5113       & 0.6240      &  0.7163     &0.4034  & 0.6092      & 0.7294      & 0.5755\\\cmidrule{2-14}
          & Gemini-2.5-Pro & 0.6888      & 0.7795      &  0.4482     & 0.7475      & 0.7977      & 0.5743      & 0.6197     &0.7313       &0.4097  & 0.6006      & 0.7601      & 0.5531\\\cmidrule{2-14}
          & Llama-3.1-8B-Instruct & 0.4912      & 0.6511      & 0.3899         & 0.5123      & 0.7345      & 0.6435&   0.5349    & 0.6277      &0.4352    &0.3443       &  0.6204     &0.3445 \\\cmidrule{2-14}
          & Mistral-7B-Instruct-v0.2 & 0.4539      &  0.6465&  0.2649     & 0.5713& 0.6697      &0.4924       & 0.3112      &0.5347       & 0.1235 & 0.2936      &0.5875       &0.1653 \\\cmidrule{2-14}
          & \textbf{CIA (Ours)} &   \hl{0.8227}    &   \hl{0.7931}    & \hl{0.7458} & \hl{0.9873}      & \hl{0.9035}      &\hl{0.8330}     &    \hl{0.9761}   &   \hl{0.9106}    &    \hl{0.8632}     & \hl{0.8699}     &\hl{0.8126}       &\hl{0.7153} \\
   
    \bottomrule
    \end{tabular}%
  }
  \caption{Comparison of inference attack performance between CIA and baselines.}
  \label{tab:main_results}
\end{table*}

\begin{table}[htbp]
    \centering
    \resizebox{\linewidth}{!}{
        \begin{tabular}{lcccccc}
            \toprule
            \multirow{2}{*}{\textbf{Dataset}} & \multicolumn{2}{c}{\textbf{G-Designer}} & \multicolumn{2}{c}{\textbf{AGP}} & \multicolumn{2}{c}{\textbf{ARG-Designer}} \\
            \cmidrule(lr){2-3} \cmidrule(lr){4-5} \cmidrule(lr){6-7}
             & \textbf{$N_{avg}$} & \textbf{$E_{avg}$} & \textbf{$N_{avg}$} & \textbf{$E_{avg}$} & \textbf{$N_{avg}$} & \textbf{$E_{avg}$} \\
            \midrule
            MMLU  & 7.0 & 8.99 & 6.0 & 10.87 & 5.42 & 7.84 \\
            GSM8K   & 5.0 & 8.19 & 5.0 & 8.45 & 3.07 & 3.14 \\
            SVAMP    & 5.0 & 8.15 & 5.0 & 8.41 & 3.05 & 3.10 \\
            HumanEval & 6.0 & 11.38 & 6.0 & 11.54 & 4.24 & 5.49 \\
            \bottomrule
        \end{tabular}
    }
    \caption{Statistical details of generated communication topologies. We report the average number of nodes ($N_{avg}$) and edges ($E_{avg}$) for each setting.}
    \label{tab:statistic_data}
\end{table}

\section{Experiments}

\subsection{Experiment Setups}
\paragraph{MAS Frameworks.} As introduced in Section~\ref{Related_Works}, generative optimization strategies for communication topologies achieve SOTA performances. Unlike heuristic methods, these strategies often require substantial resources to design communication topologies carefully, making them more valuable targets for inferring.
Accordingly, to evaluate CIA’s performance, we construct communication topologies using three well-performing generative optimization strategies: G-Designer \cite{zhang2024g}, AGP \cite{li2025adaptive}, and ARG-Designer \cite{li2025assemble}. More details about these strategies are provided in Appendix~\ref{app:algorithms}.

\paragraph{Task Datasets.}
To provide the tasks required for the topology optimization, we employ four datasets across three domains: \ding{182} General Reasoning: MMLU \cite{hendrycks2020measuring}; \ding{183} Mathematical Reasoning: GSM8K \cite{cobbe2021training}, SVAMP \cite{patel2021nlp}; and \ding{184} Code Generation: HumanEval \cite{chen2021evaluating}. For each dataset, we select 100 tasks for evaluation. More details about the datasets are provided in Appendix~\ref{app:domain}.

\paragraph{Baselines.} 
We select two closed-source LLMs (GPT-5 and Gemini-2.5-Pro) and two open-source LLMs (Llama-3.1-8B-Instruct and Mistral-7B-Instruct-v0.2) as our baseline attacks for inferring the communication topology. Specifically, we prompt them to assign confidence scores to all agent pairs for inferring the communication. We then apply a threshold of 0.5 to these scores to determine the predicted edges for evaluation. The exact prompt template is provided in Appendix~\ref{app:prompts Baselines}.

\paragraph{Metrics.}
We evaluate the performance of topology inference by measuring the prediction accuracy of all possible edges within MAS. We report Area Under the ROC Curve (AUC)~\cite{hanley1982meaning}, Accuracy (ACC) and F1-score (F1)~\cite{vujovic2021classification}.

\begin{figure*}[t]

    \centering
    \begin{subfigure}[b]{0.25\linewidth}
        \centering
        \adjustbox{trim={0} {0} {0} {0}, clip}{%
            \includegraphics[width=\linewidth]{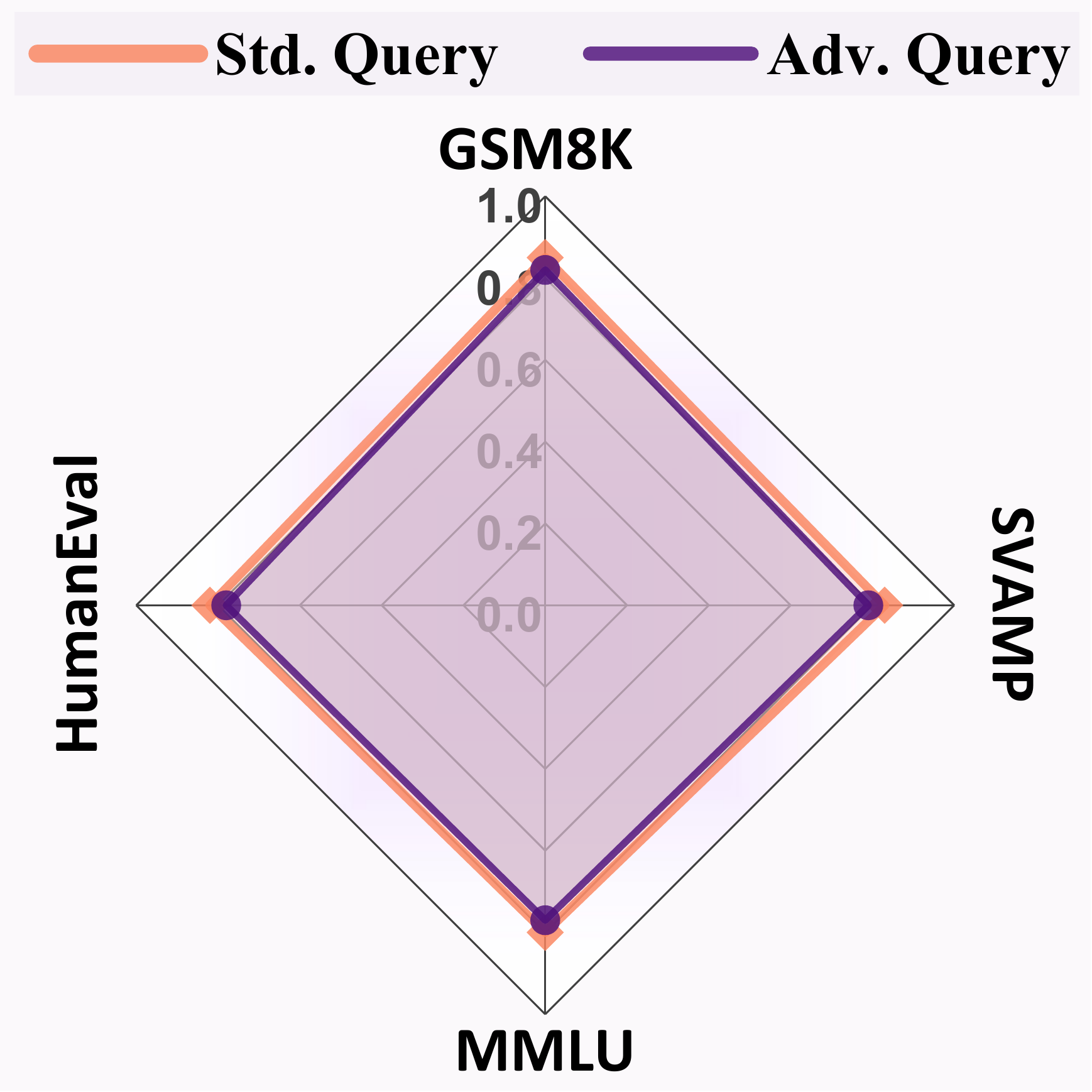}
        }
        \caption{G-Designer.}
    \end{subfigure}
    \hspace{0.5cm}
    \begin{subfigure}[b]{0.25\linewidth}
        \centering
        \adjustbox{trim={0} {0} {0} {0}, clip}{%
            \includegraphics[width=\linewidth]{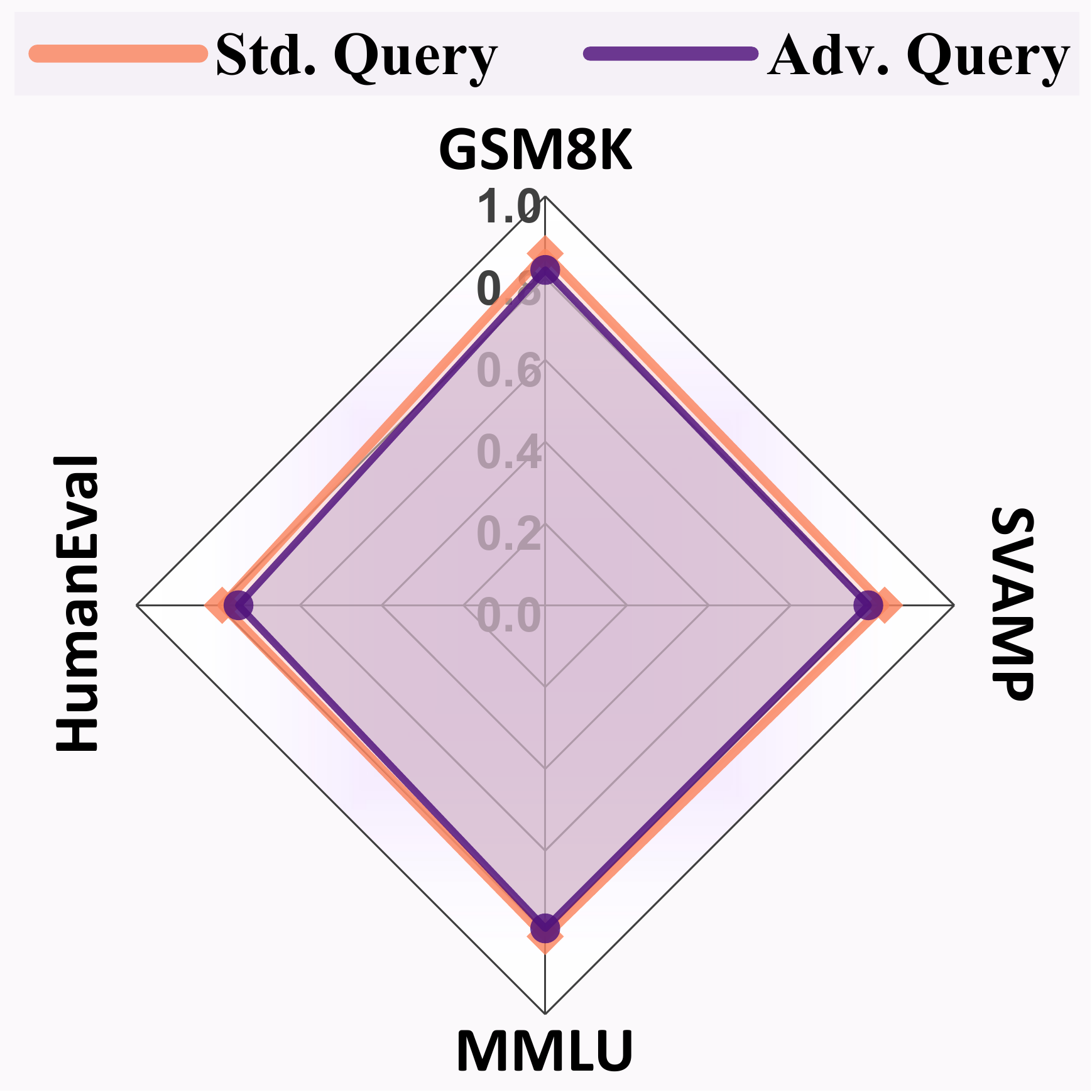}
        }
        \caption{AGP.}
    \end{subfigure}
    \hspace{0.5cm}
    \begin{subfigure}[b]{0.25\linewidth}
        \centering
        \adjustbox{trim={0} {0} {0} {0}, clip}{%
            \includegraphics[width=\linewidth]{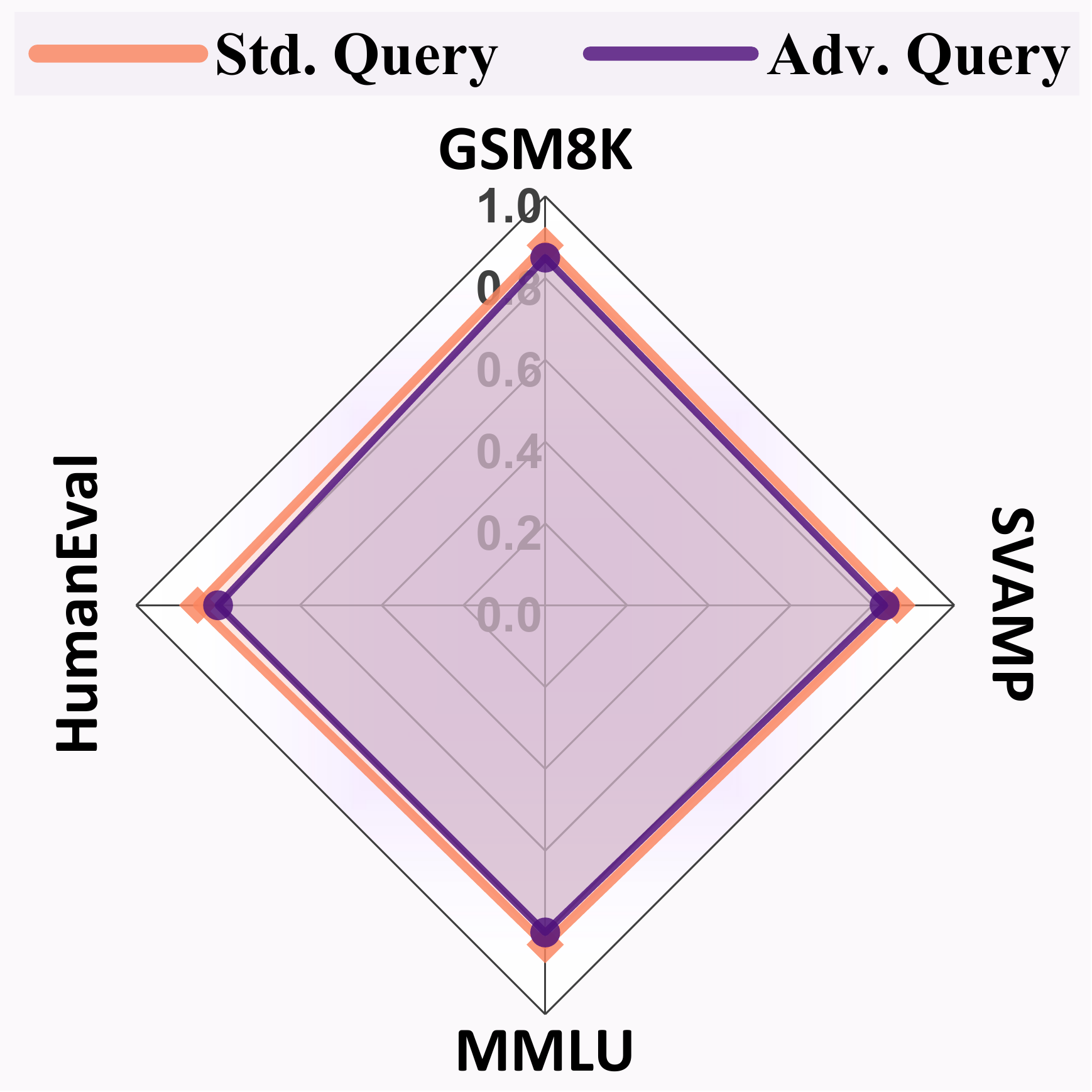}
        }
        \caption{ARG-Designer.}
    \end{subfigure}
    \caption{Comparison of MAS utility (measured by Accuracy) between \textbf{Std.Query} and \textbf{Adv.Query}.}
    \label{tab:utility}
\end{figure*}

\subsection{Inference Attack Performance}
\Cref{tab:main_results} demonstrates the communication inference results of MAS constructed by three generative optimization strategies across four datasets. The best performance is in boldface. From~\Cref{tab:main_results}, we have the following observations:

\ding{182} The communication topology can be inferred. CIA  exhibits superior performance in communication inference, with an AUC exceeding 0.75 in all cases and surpassing 0.80 in most experiments, peaking at 0.99, revealing the critical privacy risk in the communication topology of MAS.

\ding{183} A simpler communication topology is more susceptible to being inferred. As shown in~\Cref{tab:statistic_data}, ARG-Designer constructs MAS for GSM8K and SVAMP with significantly fewer average nodes and edges compared to other MAS, and our CIA achieves an AUC close to 1.0 in these cases. While having fewer nodes and edges results in lower resource consumption, it increases the risk of the communication topology leakage.

\ding{184} CIA significantly outperforms all LLM baselines. Among the baselines, closed-source models generally exhibit stronger reasoning capabilities compared to open-source models. However, all LLMs tend to assign lower confidence scores to the communication between agents, failing to effectively distinguish whether there is communication between them. 

\subsection{Effectiveness of Adversarial Query}
In this section, we evaluate our adversarial query strategy (\Cref{outputs}) by studying two primary factors: (1) \textbf{the fidelity of raw reasoning output recovery}, and (2)\textbf{ the impact of adversarial query on MAS task performance}. The latter ensures that our strategy does not degrade system functionality, which would otherwise render the inferred topologies meaningless. 

For the first factor, we use Recall (Rec) to measure the proportion of recovered agent reasoning outputs based on high semantic similarity to the ground truth, and use ROUGE-L (R-L) to evaluate the lexical precision and structural fidelity of the recovered outputs. As shown in \Cref{tab:ADQ}, our reasoning outputs induction achieves robust recovery performance across various scenarios. Notably, the effectiveness is more pronounced in MAS generated by ARG-Designer, which have simpler topologies, thus minimizing the information loss caused by multi-source aggregation during the reasoning process. thus minimizing the information loss caused by multi-source aggregation during the reasoning process.
\begin{table}[H]
    \centering
    \resizebox{\linewidth}{!}{
        \begin{tabular}{lcccccc}
            \toprule
            \multirow{2}{*}{\textbf{Dataset}} & \multicolumn{2}{c}{\textbf{G-Designer}} & \multicolumn{2}{c}{\textbf{AGP}} & \multicolumn{2}{c}{\textbf{ARG-Designer}}\\
            \cmidrule(lr){2-3} \cmidrule(lr){4-5} \cmidrule(lr){6-7} 
             & \textbf{Rec} & \textbf{R-L} & \textbf{Rec} & \textbf{R-L}  & \textbf{Rec} & \textbf{R-L}\\
            \midrule
            MMLU  & 0.90 & 0.89 & 0.89 & 0.88 & 0.93 & 0.93 \\
            GSM8K   & 0.92 & 0.91  & 0.91 & 0.91 & 0.96 & 0.95 \\
            SVAMP    & 0.92 & 0.92 & 0.93 & 0.92 & 0.95 & 0.94  \\
            HumanEval & 0.87 & 0.87 & 0.88 & 0.87& 0.93 & 0.92  \\
            \bottomrule
        \end{tabular}
    }
    \caption{Output recovery effectiveness via Adv. Query.}
    \label{tab:ADQ}
\end{table}
For the second factor, we compare the task completion accuracy between standard query (Std.Query) and our adversarial query (Adv.Query). As illustrated in \Cref{tab:utility}, the performance under the Adv.Query is nearly identical to that of Std.Query across all settings. This shows that our strategy does not degrade system performance, confirming that the inferred communication accurately reflects how MAS solve complex reasoning problems. Furthermore, by preserving collaborative integrity, CIA remains stealthy and indistinguishable from benign usage.

\subsection{Effectiveness of GBD}
\begin{figure}[t]
        \centering \includegraphics[width=0.48\textwidth]{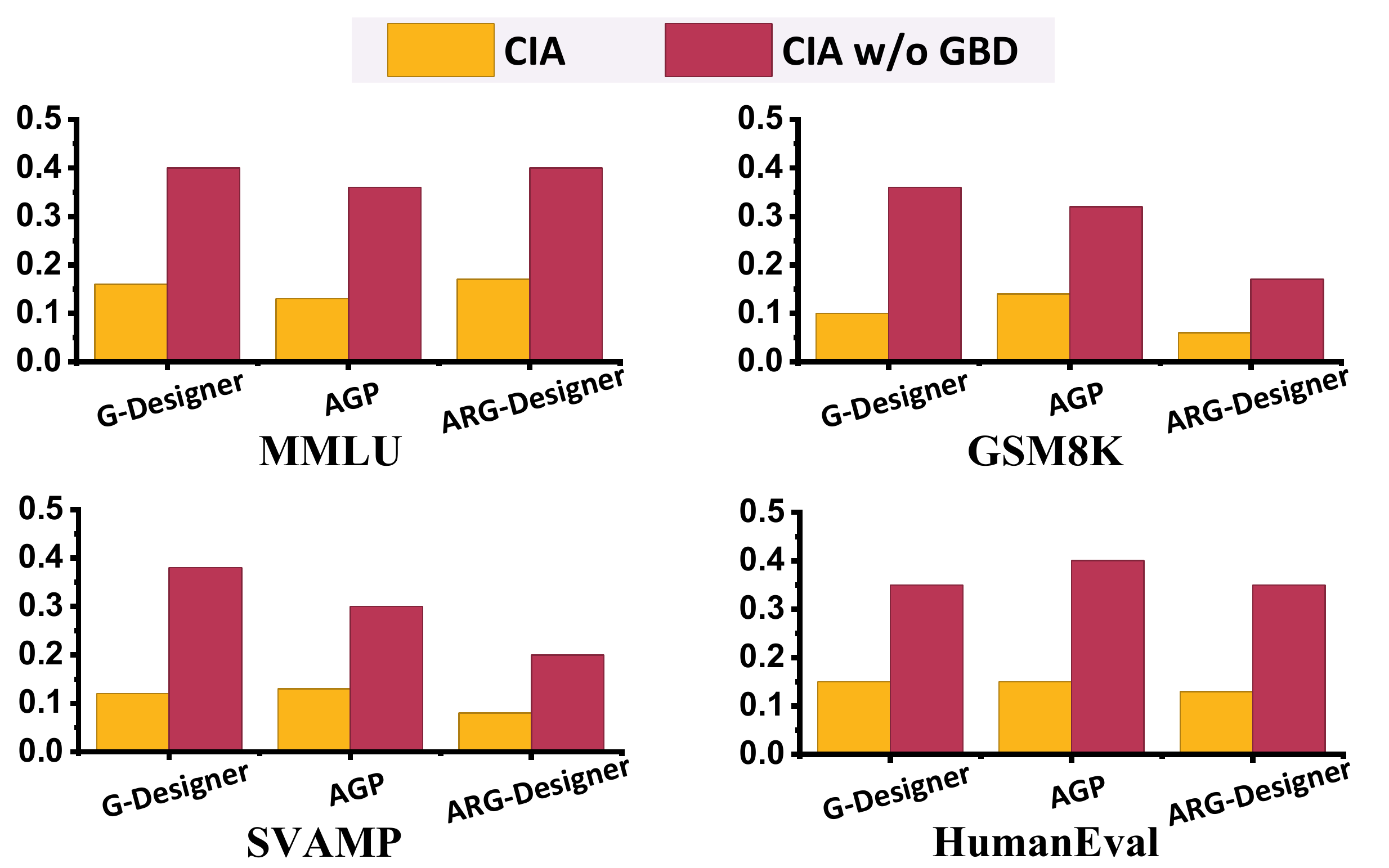}
        \caption{Impact of GBD on FPR.}
        \label{fig:fpr}
\end{figure}

In this section, we evaluate the effectiveness of GBD by comparing the False Positive Rate (FPR) and communication inference performance between CIA and CIA w/o GBD (the model variant without GBD). CIA outperforms CIA w/o GBD. As illustrated in~\Cref{fig:fpr}, by eliminating global bias, CIA substantially reduces false positives, achieving at least a 50\% reduction in FPR across all settings. Moreover, as reported in~\Cref{tab:GBD}, removing global bias mitigates the spurious correlations among agents’ reasoning outputs, leading to a significant improvement in communication inference.

\begin{table}[t]
    \centering

    \resizebox{\linewidth}{!}{
        \begin{tabular}{lcccc}
            \toprule
            {\textbf{G-Designer}} & {\textbf{MMLU}} & {\textbf{GSM8K}} &  {\textbf{SVAMP}}&  {\textbf{HumanEval}} \\
            \midrule
            CIA w/o GBD  & 0.5264 & 0.5391 & 0.5308 & 0.5115 \\
            CIA   & 0.8324 & 0.8585 & 0.8561 & 0.7594 \\\bottomrule\\ 
            \specialrule{0em}{-4pt}{-4pt}
            \toprule
            {\textbf{AGP}} & {\textbf{MMLU}} & {\textbf{GSM8K}} &  {\textbf{SVAMP}}&  {\textbf{HumanEval}} \\
            \midrule
            CIA w/o GBD  & 0.5274 & 0.5683 &0.5857 & 0.5252 \\
            CIA   & 0.8411 & 0.8804 & 0.8979 & 0.7480 \\\bottomrule\\
            \specialrule{0em}{-4pt}{-4pt}
            \toprule
            {\textbf{ARG-Designer}} & {\textbf{MMLU}} & {\textbf{GSM8K}} &  {\textbf{SVAMP}}&  {\textbf{HumanEval}} \\
            \midrule
            CIA w/o GBD  & 0.5333 & 0.6268 & 0.6294 & 0.6128  \\
            CIA   & 0.8227 & 0.9873 & 0.9761 & 0.8699 \\\bottomrule
        \end{tabular}
    }
     \caption{Impact of GBD on attack performance (AUC).}
    \label{tab:GBD}
\end{table}

\subsection{Effectiveness of LWS}
\begin{figure}[t]
        \centering
        \includegraphics[width=0.48\textwidth]{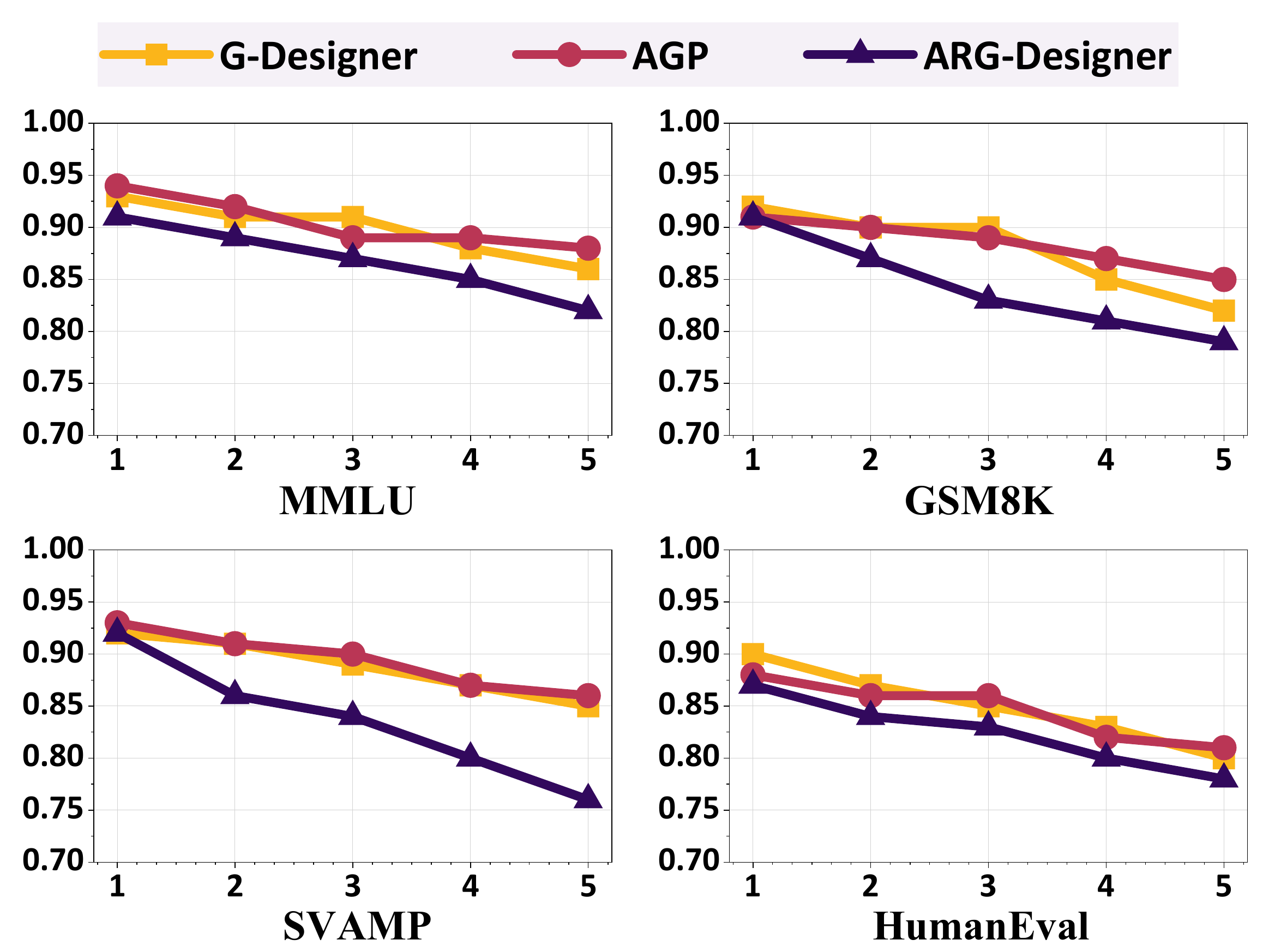}
        \caption{Precision of Top-$k$ high-confidence edges.}
        \label{fig:ws}
\end{figure}

In this section, we evaluate the effectiveness of LLM-guided Weak Supervision (LWS). Since LLMs struggle with direct full-topology inference as shown in~\Cref{tab:main_results}, we first assess the precision of Top-$k$ high-confidence edges to verify the reliability of these supervision signals. We then conduct an ablation study to compare CIA with CIA w/o LWS (the model variant without LWS). \Cref{fig:ws} demonstrates that the LLM performs well in the precision evaluation of Top-$k$ high-confidence edges, particularly where $k \le 3$, indicating that these weak supervision signals are reliable for inference. Consequently, as shown in~\Cref{tab:ws}, LWS improves AUC across all settings, validating its effectiveness in refining the debiased representations and enhancing the inference performance.

\begin{table}[t]
    \centering
    \resizebox{\linewidth}{!}{
        \begin{tabular}{lcccc}
            \toprule
            {\textbf{G-Designer}} & {\textbf{MMLU}} & {\textbf{GSM8K}} &  {\textbf{SVAMP}}&  {\textbf{HumanEval}} \\
            \midrule
            CIA w/o LWS  & 0.7856 & 0.8042 & 0.7852 & 0.7348 \\
            CIA   & 0.8324 & 0.8585 & 0.8561 & 0.7594 \\\bottomrule\\ 
            \specialrule{0em}{-4pt}{-4pt}
            \toprule
            {\textbf{AGP}} & {\textbf{MMLU}} & {\textbf{GSM8K}} &  {\textbf{SVAMP}}&  {\textbf{HumanEval}} \\
            \midrule
            CIA w/o LWS  & 0.7960 & 0.8243 &0.8471 & 0.7061  \\
            CIA   & 0.8411 & 0.8804 & 0.8979 & 0.7480 \\\bottomrule\\
            \specialrule{0em}{-4pt}{-4pt}
            \toprule
            {\textbf{ARG-Designer}} & {\textbf{MMLU}} & {\textbf{GSM8K}} &  {\textbf{SVAMP}}&  {\textbf{HumanEval}} \\
            \midrule
            CIA w/o LWS  & 0.7671 & 0.9012 & 0.8992 & 0.7724  \\
            CIA   & 0.8227 & 0.9873 & 0.9761 & 0.8699 \\\bottomrule
        \end{tabular}
    }
     \caption{Impact of LWS on attack performance (AUC).}
    \label{tab:ws}
\end{table}

\section{Conclusion}
This paper investigates the privacy risk of MAS communication topologies being inferred, which poses significant security and IP threats. We propose a restrictive black-box attack, CIA, which operates in two stages: first, it constructs adversarial queries to reveal all agent reasoning outputs; second, it infers the communication topology by analyzing the semantic correlations among agents using global bias disentanglement and LLM-guided weak supervision. Extensive experiments show that CIA can effectively infer communication topologies, highlighting the inherent privacy risk of MAS communication.

\section*{Acknowledgments}
This work is supported by the National Natural Science Foundation of China (No.62402491, No.U2336202, No.62472416) and the China Postdoctoral Science Foundation (No.2025M771524). 

\section*{Limitations}
CIA employs a recursive estimation of Total Correlation (TC) to optimize the information-theoretic objectives in Eq.~(\ref{loss_bias}), as detailed in Appendix~\ref{app:TC}. However, accurately estimating multivariate mutual information among high-dimensional vectors remains highly challenging, leaving room for improvement in our approximation strategy. Moreover, the current LLM-guided weak supervision in CIA captures only first-order topological information; exploiting higher-order topological patterns to further strengthen the attack remains an open research direction.

\label{sec:bibtex}


\bibliography{custom}

\appendix

\section {Details of Post-processing Procedures}
\label{app:post}
\begin{figure}[htbp]
        \centering
        \includegraphics[width=0.48\textwidth]{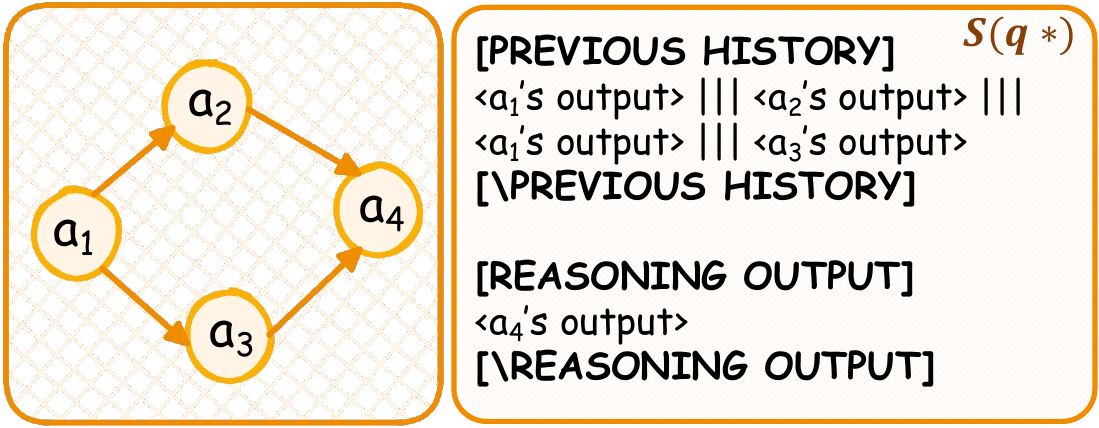}
        \caption{An illustrative example of $\mathcal{S}(q^*)$.}
        \label{fig:sq}
\end{figure}

This section details the post-processing procedures applied to $\mathcal{S}(q^{*})$. As illustrated in~\Cref{fig:sq}, $\mathcal{S}(q^{*})$ consists of two sections: [PREVIOUS HISTORY] and [REASONING OUTPUT]. To format $\mathcal{S}(q^*)$ into the structured list $\mathcal{R}^*$ for downstream communication inference, firstly, we partition the [PREVIOUS HISTORY] section of $\mathcal{S}(q^{*})$ using the delimiter “|||” to extract reasoning outputs from intermediate agents. Next, since a predecessor's output may be inherited by multiple subsequent agents, we apply backward deduplication to the partitioned results. Finally, since the [PREVIOUS HISTORY] section in $\mathcal{S}(q^{*})$ doesn't record the output of the decision agent, we append the content in the [REASONING OUTPUT] section of $\mathcal{S}(q^{*})$.

\section{Computational Details of Eq.(\ref{loss_bias})}
\label{app:TC}
In this section, we detail the computation process of Eq.(\ref{loss_bias}). The equation consists of two terms:
(i) the negative multivariate mutual information among all outputs' bias representations: $- \ \mathcal{I}(\mathbf{z}_1^b; \dots; \mathbf{z}_n^b)$, and (ii) the sum of mutual information between the debiased and bias representations for each agent's output: $\sum_{i=1}^n\ \mathcal{I}(\mathbf{z}_i^d; \mathbf{z}_i^b)$.

For the first term, we introduce Total Correlation to estimate the multivariate mutual information. First, we provide the definitions of mutual information (MI)~\cite{belghazi2018mutual} and Total Correlation (TC)~\cite{watanabe1960information}. 

Given two random variables $\bm{x}$ and $\bm{y}$, their MI is defined as
\begin{equation}
    \mathcal{I}(\bm{x}; \bm{y}) = \mathbb{E}_{p(\bm{x}, \bm{y})} \left[ \log \frac{p(\bm{x}, \bm{y})}{p(\bm{x})p(\bm{y})} \right]
\end{equation}

For multi-variable scenarios, TC is defined as
\begin{equation}
\begin{split}
    \mathcal{TC}(\bm{X}) &= \mathcal{TC}(\bm{x}_1, \bm{x}_2, \dots, \bm{x}_n) \\
    &= \mathbb{E}_{p(\bm{x}_1, \dots, \bm{x}_n)} \left[ \log \frac{p(\bm{x}_1, \dots, \bm{x}_n)}{p(\bm{x}_1) \dots p(\bm{x}_n)} \right].
\end{split}
\end{equation}

Based on the above definitions, the connection between TC and MI is described by the following theorem: 
\begin{theorem}[\citealp{bai2023estimating}]
Let $\bm{X} = (\bm{x_1}, \bm{x_2}, \dots, \bm{x_n})$ be a group of random variables. Suppose set $\mathcal{A} = \{i_1, i_2, \dots, i_k\} \subseteq \{1, 2, \dots, n\}$ is an index subgroup. $\bar{\mathcal{A}} = \{ i : i \notin \mathcal{A} \}$ is the complementary set of $\mathcal{A} $. Denote $\bm{X}_\mathcal{A} = (x_{i_1}, x_{i_2}, \dots, x_{i_k})$ as the selected variables from $\bm{X}$ with the indexes in $\mathcal{A}$. Then we have $
\mathcal{TC}(\bm{X}) = \mathcal{TC}(\bm{X}_\mathcal{A}) + \mathcal{TC}(\bm{X}_{\bar{\mathcal{A}}}) + \mathcal{I}(\bm{X}_\mathcal{A}; \bm{X}_{\bar{\mathcal{A}}}).$
\label{theorem}
\end{theorem}
\Cref{theorem} reveals that TC can be equivalently decomposed into the internal correlations within subgroups and the MI between these subgroups. Based on this property, we can recursively represent the TC of the subgroups in terms of MI terms for lower-level subgroups. Consequently, by iteratively partitioning the set into a preceding subset $\bm{X}_{1:i}$ and the current variable $\bm{x}_{i+1}$, TC can be formulated as a summation of progressive MI terms:
\begin{equation}
\mathcal{TC}(\bm{X}) = \sum_{i=1}^{n-1} \mathcal{I}(\bm{X}_{1:i}; \bm{x}_{i+1}),
\end{equation}

Based on this, our multivariate mutual information can be reformulated as
\begin{equation}
    \mathcal{I}(\mathbf{z}_1^b; \dots; \mathbf{z}_n^b) = \mathcal{TC}(\mathbf{Z}^b) = \sum_{i=1}^{n-1} \mathcal{I}(\mathbf{Z}_{1:i}^b; \mathbf{z}_{i+1}^b),
\label{eq:tc_summation}
\end{equation}
where $\mathbf{Z}^b=(\mathbf{z}_1^b, \dots, \mathbf{z}_n^b)$, and $\mathbf{Z}_{1:i}^b=(\mathbf{z}_1^b, \dots, \mathbf{z}_i^b)$ denote a subset of variables with indexes from $1$ to $i$. 

To estimate each MI term, we perturb each task multiple times to elicit diverse responses and use the collected outputs to form an empirical approximation of each agent’s output distribution. We then apply the InfoNCE~\cite{oord2018representation} to estimate the MI terms in the summation:
\begin{equation}
\begin{split}
\mathcal{I}(\mathbf{Z}_{1:i}^b; \mathbf{z}_{i+1}^b)
= \mathbb{E}\Bigg[
\frac{1}{M}\sum_{p=1}^{M}
\Big(
\phi(\mathbf{Z}_{1:i,p}^b, \mathbf{z}_{i+1,p}^b)
\\
-\log\Big(
\frac{1}{M}\sum_{q=1}^{M}
\exp\big(\phi(\mathbf{Z}_{1:i,p}^b, \mathbf{z}_{i+1,q}^b)\big)
\Big)
\Big)
\Bigg],
\end{split}
\label{eq:infonce}
\end{equation}

For the second term in Eq.(\ref{loss_bias}), we also use InfoNCE to estimate the MI between the debiased and bias representations for each agent’s output:
\begin{equation}
\begin{split}
\mathcal{I}(\mathbf{z}_i^d; \mathbf{z}_i^b)
= \mathbb{E}\Bigg[
\frac{1}{M}\sum_{p=1}^{M}
\Big(
\phi(\mathbf{z}_{i,p}^d, \mathbf{z}_{i,p}^b)
\\
-\log\Big(
\frac{1}{M}\sum_{q=1}^{M}
\exp\big(\phi(\mathbf{z}_{i,p}^d, \mathbf{z}_{i,q}^b)\big)
\Big)
\Big)
\Bigg].
\end{split}
\label{eq:infonce_zd_zb}
\end{equation}


\section{Formulations of $\mathcal{L}_{\mathrm{pos}}$ and $\mathcal{L}_{\mathrm{neg}}$}
\label{pos}
Here we first give the formulation of  $\mathcal{L}_{\mathrm{pos}}$:
\begin{equation}
\begin{aligned}
\mathcal{L}_{\mathrm{pos}}(a_i,a_j)
&= (1-\alpha) \cdot \log(\mathrm{Sim}(\mathbf{z}_i^d, \mathbf{z}_j^d)) \\\
&\quad + \alpha \cdot \log(1 - \mathrm{Sim}(\mathbf{z}_i^d, \mathbf{z}_j^d)),
\end{aligned}
\label{eq:pos}
\end{equation}
where $\alpha$ is the label-smoothing coefficient, $\mathrm{Sim}(\cdot,\cdot)$ is a distance-based similarity function, and $\mathbf{z}_i^d$ and $\mathbf{z}_j^d$ denote the debiased representations corresponding to the outputs of agents $a_i$ and $a_j$. Accordingly, the formulation of $\mathcal{L}_{\mathrm{neg}}$ is
 
\begin{equation}
\begin{aligned}
\mathcal{L}_{\mathrm{neg}}(a_u,a_v)
&=  (1-\alpha) \cdot \log(1 - \mathrm{Sim}(\mathbf{z}_u^d, \mathbf{z}_v^d))\\\
&\quad +\alpha \cdot \log(\mathrm{Sim}(\mathbf{z}_u^d, \mathbf{z}_v^d)) .
\end{aligned}
\label{eq:neg}
\end{equation}

\section{Generative Optimization Strategies}
\label{app:algorithms}
The generative optimization strategies for the communication topology used in our experiments are introduced below.
\begin{itemize}
    \item G-Designer~\cite{zhang2024g} is a topology optimization framework that learns effective multi-agent communication topologies by modeling agent interactions as a graph and optimizing connectivity to improve collaborative reasoning performance.

    \item AGP \cite{li2025adaptive} proposes an adaptive graph pruning strategy that iteratively removes redundant or ineffective communication links, resulting in more efficient and task-relevant multi-agent interaction topologies.

    \item ARG-Designer \cite{li2025assemble} reframes multi-agent system design as conditional autoregressive graph generation. By jointly optimizing agent composition and topology, it constructs customized topologies from scratch to enable task-adaptive coordination.
\end{itemize}

\section{Task Datasets}
\label{app:domain}
The task datasets used in our experiments are introduced below.
\begin{itemize}
    \item MMLU~\cite{hendrycks2020measuring} is a benchmark designed to evaluate general reasoning and knowledge understanding across diverse subject domains, covering a wide range of factual, logical, and conceptual questions.

    \item GSM8K~\cite{cobbe2021training} focuses on complex, multi-step mathematical reasoning, requiring models to perform precise arithmetic operations and logical deductions to solve diverse, grade-school–level word problems.

    \item SVAMP~\cite{patel2021nlp} targets robustness in mathematical reasoning by introducing systematic structural and linguistic variations to math problems, testing whether models truly understand complex problem semantics rather than relying on spurious surface cues.

    \item HumanEval~\cite{chen2021evaluating} is a code generation benchmark that assesses a model’s ability to synthesize correct and executable programs from natural language specifications.
\end{itemize}


\section{Model Variant for Disentanglement}
In this section, we introduce a model variant, termed CIA-Sub, which approaches the distanglement of global bias in a different way. Instead of using two separate encoders to learn the debiased representations and the bias representations, CIA-Sub uses a single encoder to obtain the bias representations, while the debiased representations are defined by subtracting the bias representations from the initial representations:
\begin{equation}
\mathbf{z}_i^b = E^b(\mathbf{h}_i),\ \ \mathbf{z}_i^d = \mathbf{h}_i - \mathbf{z}_i^b.
\end{equation}
All loss components for CIA-Sub remain unchanged from those of CIA.

As shown in~\Cref{tab:sub}, CIA performs better than CIA-Sub. We suppose the reason is that in CIA-Sub, $\mathbf{z}_i^d$ is entirely dependent on the quality of $\mathbf{z}_i^b$, while in CIA, using two separate encoders allows the debiased representations to be explicitly refined to capture useful information relevant to the communication.

\begin{table}[H]
    \centering
    \resizebox{\linewidth}{!}{
        \begin{tabular}{lcccc}
            \toprule
            {\textbf{G-Designer}} & {\textbf{MMLU}} & {\textbf{GSM8K}} &  {\textbf{SVAMP}}&  {\textbf{HumanEval}} \\
            \midrule
            CIA-Sub  & 0.7432 & 0.7689 & 0.7455 & 0.6509 \\
            CIA   & 0.8324 & 0.8585 & 0.8561 & 0.7594 \\\bottomrule\\ 
            \specialrule{0em}{-4pt}{-4pt}
            \toprule
            {\textbf{AGP}} & {\textbf{MMLU}} & {\textbf{GSM8K}} &  {\textbf{SVAMP}}&  {\textbf{HumanEval}} \\
            \midrule
            CIA-Sub  & 0.7181 & 0.7625 &0.7964 & 0.6622  \\
            CIA   & 0.8411 & 0.8804 & 0.8979 & 0.7480 \\\bottomrule\\
            \specialrule{0em}{-4pt}{-4pt}
            \toprule
            {\textbf{ARG-Designer}} & {\textbf{MMLU}} & {\textbf{GSM8K}} &  {\textbf{SVAMP}}&  {\textbf{HumanEval}} \\
            \midrule
            CIA-Sub  & 0.7454 & 0.8654 & 0.8294 & 0.7669  \\
            CIA   & 0.8227 & 0.9873 & 0.9761 & 0.8699 \\\bottomrule
        \end{tabular}
    }
     \caption{AUC comparison between CIA-Sub and CIA.}
    \label{tab:sub}
\end{table}

\section{Case studies}
In this section, we present case studies for communication inference. Specifically, we visualize the similarity matrices and the inferred communication topologies produced by CIA w/o GBD and CIA, and compare them against the ground-truth adjacency matrix and communication topology. Notably, since the similarity matrix is symmetric, we symmetrize the ground-truth adjacency matrix for a more intuitive comparison.

\begin{figure}[t]
    \centering
    \begin{subfigure}[t]{\columnwidth}
        \centering
        \includegraphics[width=0.4\columnwidth]{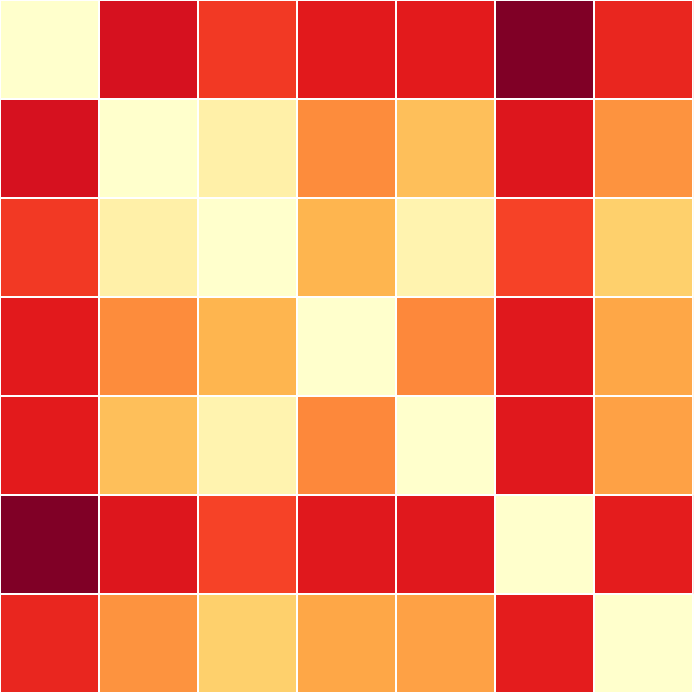}
        \includegraphics[width=0.4\columnwidth]{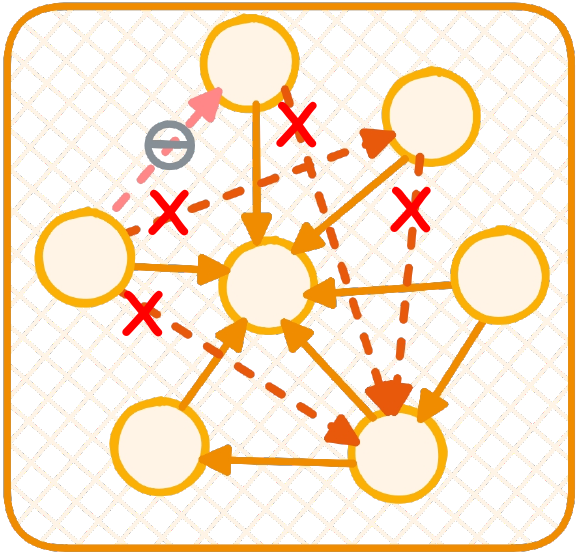}
        \caption{CIA w/o GBD.}
        \label{fig:row1}
    \end{subfigure}
    \begin{subfigure}[t]{\columnwidth}
        \centering
        \includegraphics[width=0.4\columnwidth]{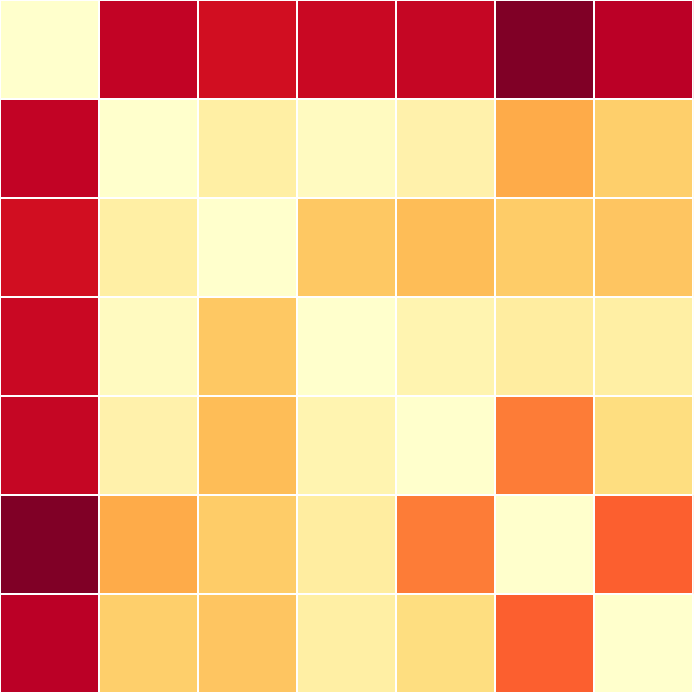}
        \includegraphics[width=0.4\columnwidth]{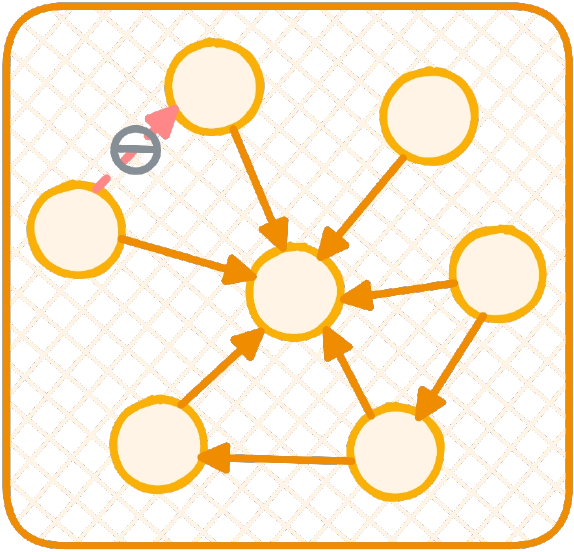}
        \caption{CIA.}
        \label{fig:row2}
    \end{subfigure}

    \begin{subfigure}[t]{\columnwidth}
        \centering
        \includegraphics[width=0.4\columnwidth]{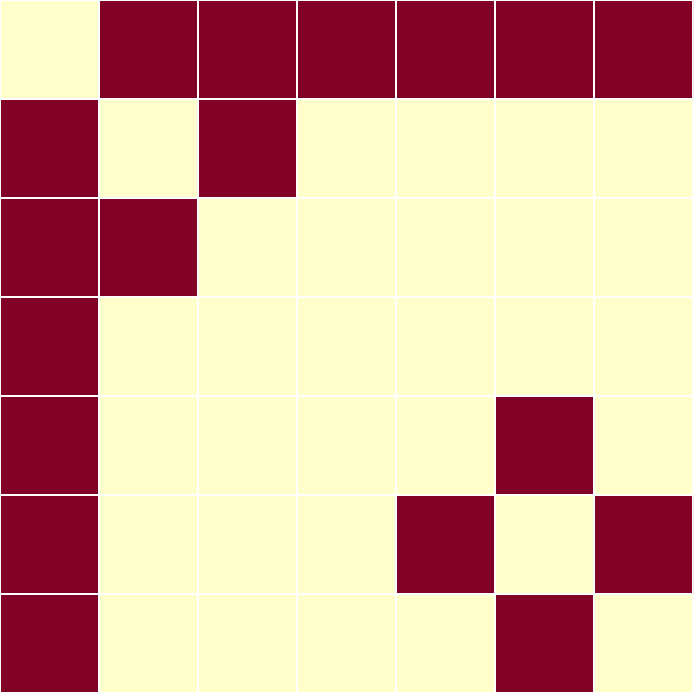}
        \includegraphics[width=0.4\columnwidth]{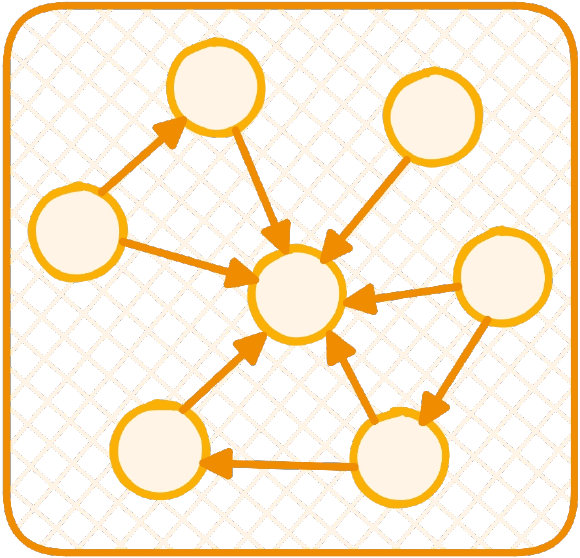}
        \caption{Ground-truth.}
        \label{fig:row3}
    \end{subfigure}
    \caption{A case study of the communication topology generated by G-Designer on MMLU.}
    \label{fig:case_1}
\end{figure}

\begin{figure}[H]
    \centering
    \begin{subfigure}[t]{\columnwidth}
        \centering
        \includegraphics[width=0.4\columnwidth]{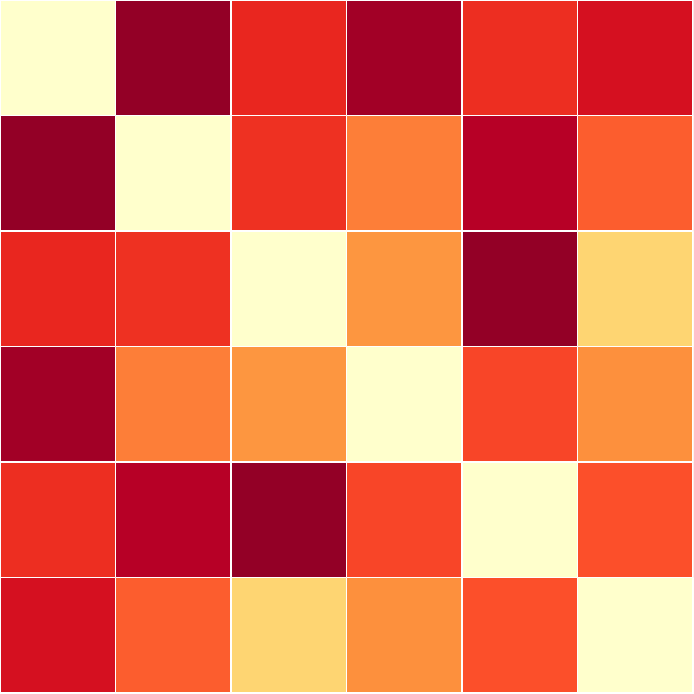}
        \includegraphics[width=0.4\columnwidth]{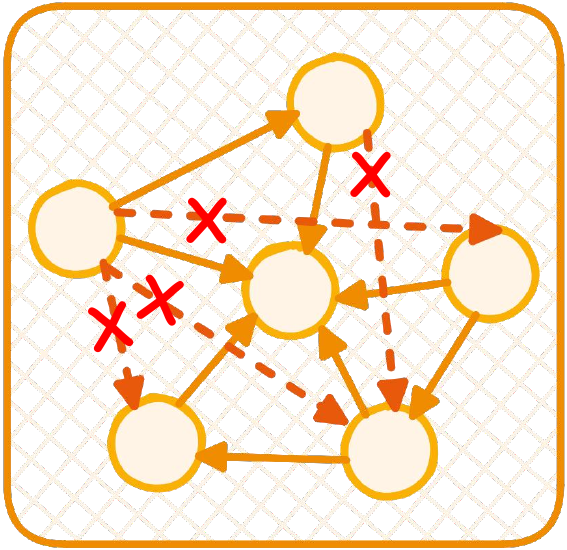}
        \caption{CIA w/o GBD.}
        \label{fig:row4}
    \end{subfigure}
    \begin{subfigure}[t]{\columnwidth}
        \centering
        \includegraphics[width=0.4\columnwidth]{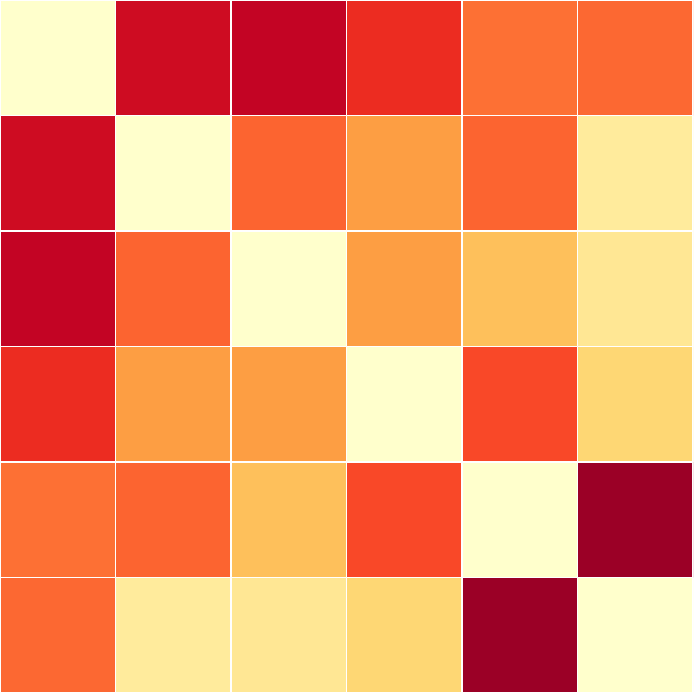}
        \includegraphics[width=0.4\columnwidth]{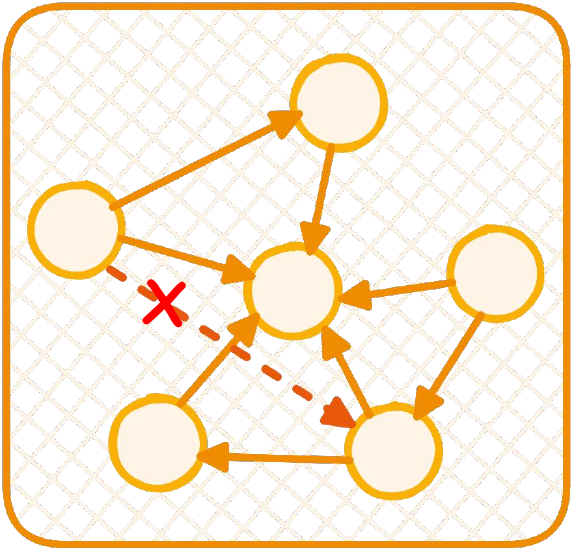}
        \caption{CIA.}
        \label{fig:row5}
    \end{subfigure}

    \begin{subfigure}[t]{\columnwidth}
        \centering
        \includegraphics[width=0.4\columnwidth]{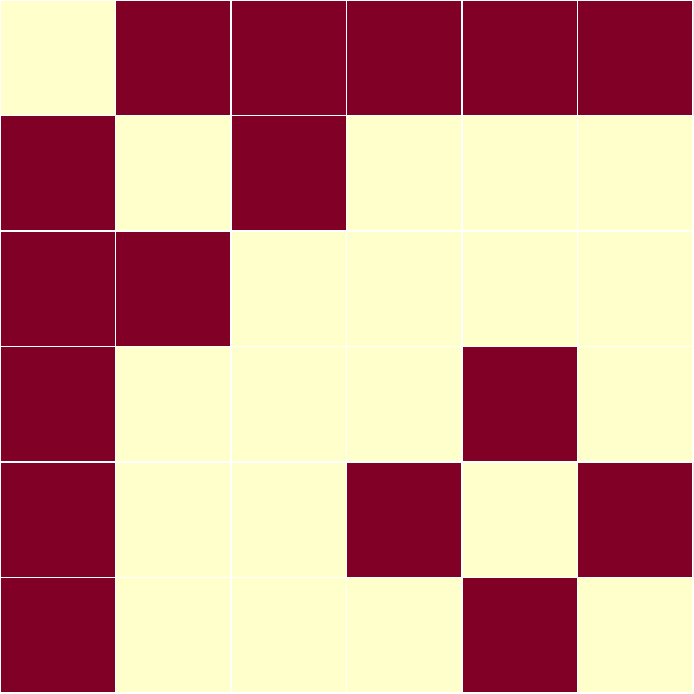}
        \includegraphics[width=0.4\columnwidth]{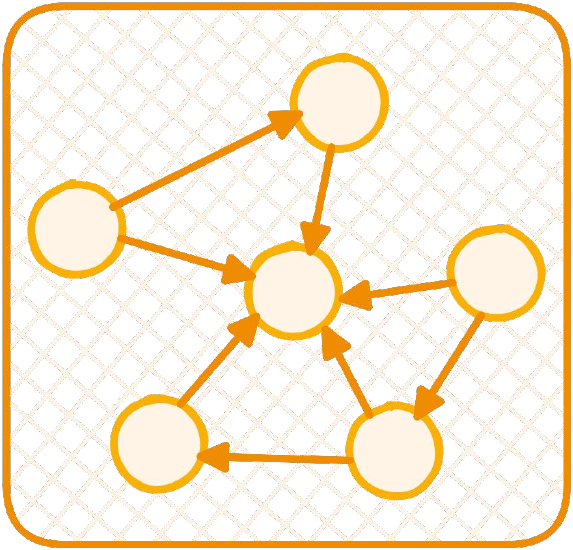}
        \caption{Ground-truth.}
        \label{fig:row6}
    \end{subfigure}
    \caption{A case study of the communication topology generated by AGP on MMLU.}
    \label{fig:case_2}
\end{figure}

\begin{figure}[H]
    \centering
    \begin{subfigure}[t]{\columnwidth}
        \centering
        \includegraphics[width=0.4\columnwidth]{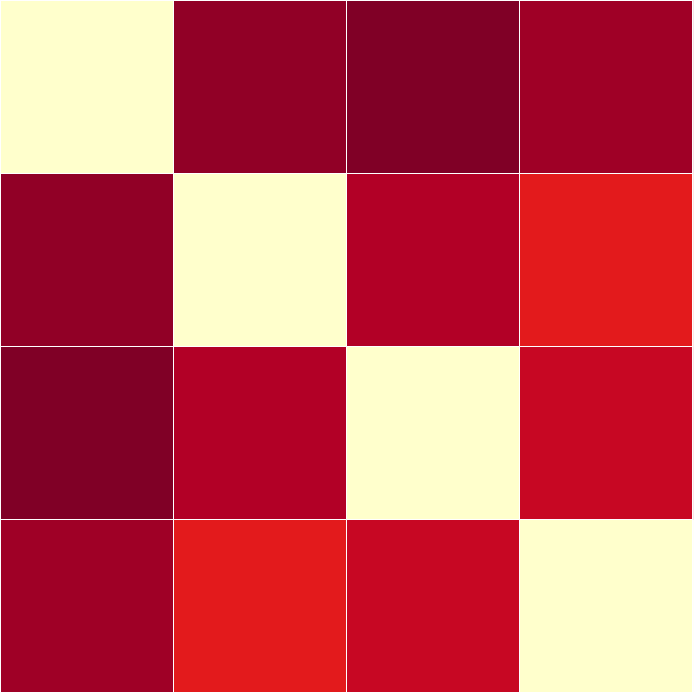}
        \includegraphics[width=0.4\columnwidth]{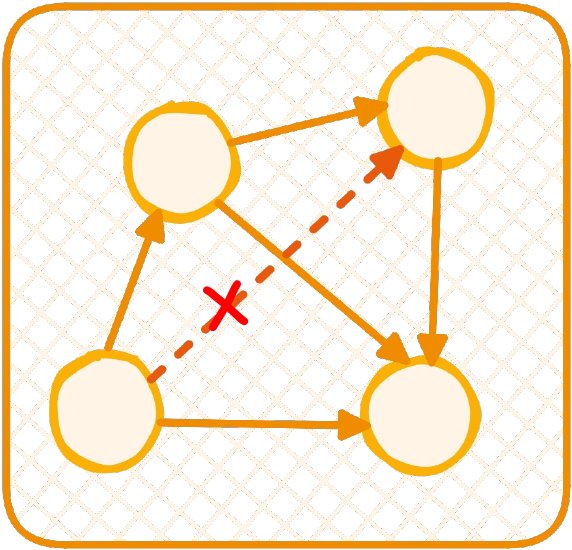}
        \caption{CIA w/o GBD.}
        \label{fig:row7}
    \end{subfigure}

    \begin{subfigure}[t]{\columnwidth}
        \centering
        \includegraphics[width=0.4\columnwidth]{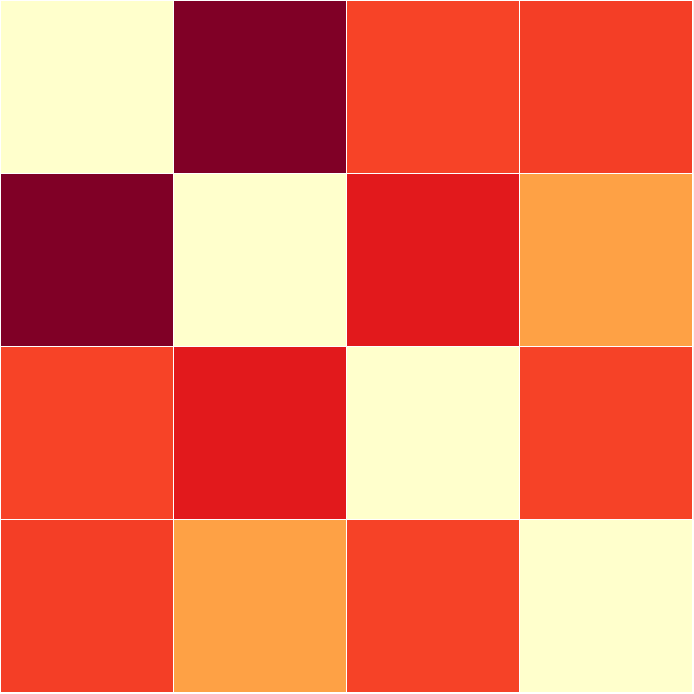}
        \includegraphics[width=0.4\columnwidth]{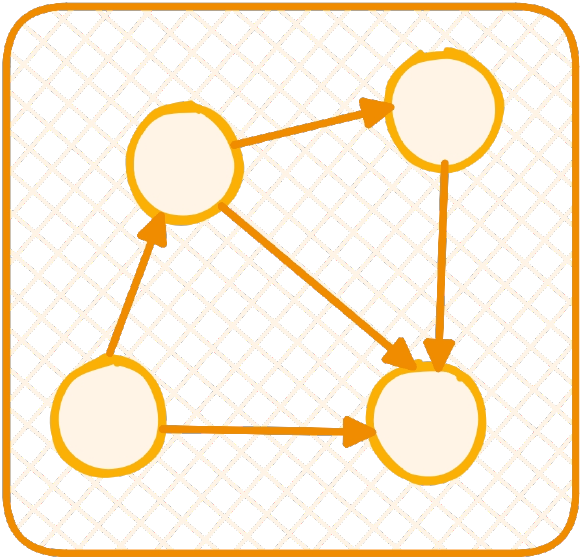}
        \caption{CIA.}
        \label{fig:row8}
    \end{subfigure}
    \begin{subfigure}[t]{\columnwidth}
        \centering
        \includegraphics[width=0.4\columnwidth]{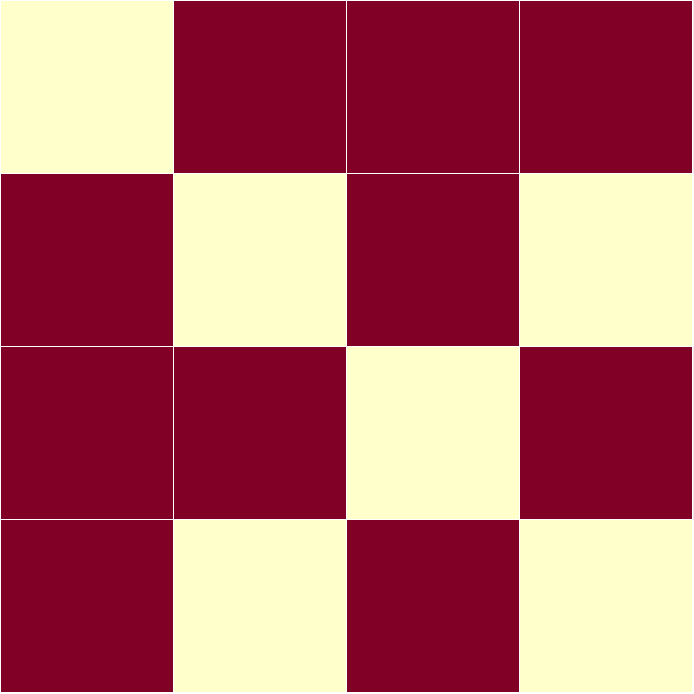}
        \includegraphics[width=0.4\columnwidth]{fig/case_2_1.pdf}
        \caption{Ground-truth.}
        \label{fig:row9}
    \end{subfigure}

    \caption{A case study of the communication topology generated by ARG-Dsigner on MMLU.}
    \label{fig:case_3}
\end{figure}

\Cref{fig:case_1,fig:case_2,fig:case_3} present three case studies, where the communication topologies are generated by G-Designer, AGP, and ARG-Designer on MMLU, respectively. These case studies visualize the spurious correlations induced by global bias and their impact on communication inference, thereby demonstrating the effectiveness of our GBD.

\section{Implementation Details}
For our implementation, we utilize all-MiniLM-L6-v2 as the pretrained language model $f_\theta$ to generate the initial representations. The dimensions of both the debiased and biased representations in~\nameref{para:gbd} are set to 768, which is twice the native output dimension of $f_\theta$. We employ GPT-5 as the teacher LLM to provide weak supervision signals in~\nameref{para:ws}. The threshold $\tau$ in Eq.(\ref{eq:link}) is set to 0.5, following common practice~\cite{zhou2022ood}. Additionally, the label-smoothing coefficient $\alpha$ in Eq.(\ref{eq:pos}) and Eq.(\ref{eq:neg}) is set to 0.1, consistent with the practice in \citet{dettmers2018convolutional}.

\section{Hyperparameter Analysis}
\begin{figure}[htbp]
        \centering \includegraphics[width=0.48\textwidth]{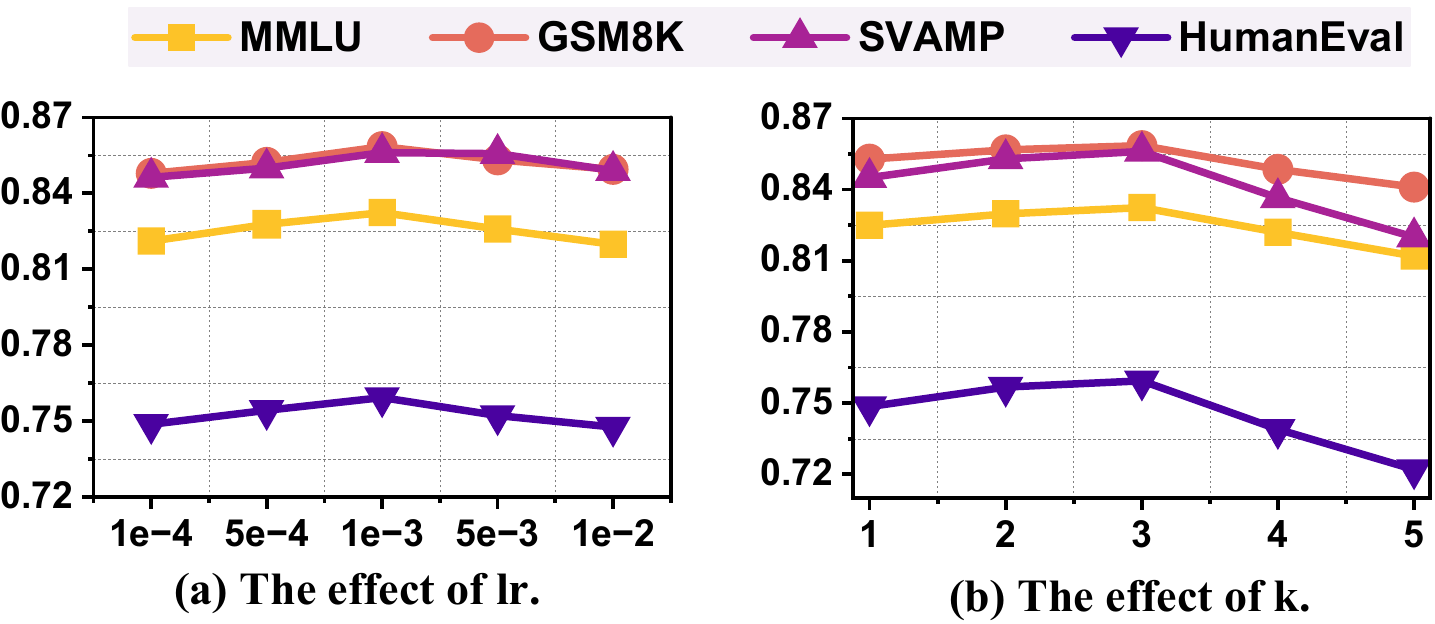}
        \caption{Hyperparameter analysis of CIA.}
        \label{fig:super}
\end{figure}

We conduct a grid search to select the optimal hyperparameter values. Specifically, we focus on tuning two primary parameters: the learning rate $\rm lr$ and $k$, the number of the highest confidence scores edges returned by the teacher LLM in LWS. The search intervals for these parameters are $\{1e-4, 5e-4, 1e-3, 5e-3, 1e-2\}$ and $\{1, 2, 3, 4, 5\}$, respectively. We conduct these hyperparameter experiments on MAS generated by G-Designer. The corresponding results are illustrated in~\Cref{fig:super}. From it, we have the following conclusions:

\ding{182} CIA performs best at a learning rate of 1e-3, so appropriately increasing the learning rate can improve the CIA's inference quality. A smaller learning rate would result in slow convergence and inadequate learning, while a larger learning rate may cause gradient oscillations, slightly degrading performance.

\ding{183} CIA achieves the best performance when $k=3$. A smaller $k$ makes the debiased representations difficult to capture sufficient topological information, while for a large $k$, the teacher LLM's precision decreases more noticeably, as shown in~\Cref{fig:ws}, introducing more incorrect edges, which brings noise and misleads $\mathbf{z}_i^d$ 
to learn topological information deviating from the true topology, thereby degrading the performance.

\section{Prompts for LWS}
\label{app:prompts Weak Supervision}

\Cref{tab:prompt_ws} presents the exact prompts used to instruct teacher LLM to return the top-$k$ communication edges with the highest confidence scores.

\begin{table*}[t]
\centering
\begin{tcolorbox}[
    enhanced,
    width=\textwidth,
    boxrule=0.6pt,
    arc=2mm,
    left=4pt,right=4pt,top=4pt,bottom=4pt,
    title=\textbf{Prompt for LLM Inferring Top-$k$ Edge},
    coltitle=black,
    colback=teal!1!white,
    colframe=teal!50!white,
    boxed title style={
      colback=teal!12!white,
      colframe=teal!28!black,
      boxrule=0.5pt,
      arc=2mm
    },
]
\textbf{\textit{System Prompt}:}

\textbf{Role}: You are an expert in Multi-Agent System topology analysis. 
Your core capability is Topology Inference: infer the hidden communication topology by analyzing textual interaction logs.

\textbf{Task}: Analyze the provided agent reasoning outputs to identify communication between agents. You must quantify the likelihood of a directed edge existing from Agent $i$ to Agent $j$ (Edge $[i \to j]$) using a Confidence Score (0-100\%).

Confidence Scoring Framework: Assign a score based on the strength of the evidence found in Agent $j$'s response relative to Agent $i$'s output:

\textbf{High Confidence (80-100\%): Explicit Reference \& Strict Execution.}

Explicit Citation: Agent $j$ explicitly mentions Agent $i$ (e.g., "Using the code provided above...", "Following the previously suggested approach...").

Hard Dependency: Agent $j$ executes code, runs a SQL query, or parses a specific data topology that was strictly defined or generated by Agent $i$. Agent $j$ cannot function without this specific asset.

\textbf{Medium Confidence (50-79\%): Content Dependency \& Semantic Alignment.}

Unique Information Flow: Agent $j$ solves a problem using unique information or a specific plan provided only by Agent $i$, even without explicit citation.

High Semantic Similarity: Agent $j$'s content exhibits high semantic overlap with Agent $i$. This includes reusing specific terminologies, variable names, or continuing a niche topic introduced by $i$, indicating a direct continuation of the context.

Logical Mapping: There is a clear "Question $\to$ Answer" or "Task $\to$ Solution" mapping between $i$ and $j$.

\textbf{Low Confidence (<50\%): Weak Logical Sequence.}

Agent $j$ and Agent $i$ have a weak logical sequence, the content is generic, or the solution could have been derived independently without Agent $i$'s specific input.

\textbf{Important}:

1. The MAS communication topology is directed, and edge direction must follow the temporal order in the provided textual interaction logs. 
Specifically, if Agent i's output appears earlier than Agent j's output in the logs, you may predict an edge [i -> j] but you must NOT predict [j -> i]. 
In other words, information can only flow forward in the log order (earlier agent -> later agent).

2. ID Consistency: Use the exact Agent IDs (e.g., 0, 1, 2) provided in the input.

3. Completeness: Output the Top $k$ Highest Confidence Edges.

4. Confidence Sorting: Output the edges sorted by Confidence in descending order, starting with the highest confidence edge.

5. No Markdown: Do not use markdown code blocks.

Provide a raw JSON list in the following format. Do not include any introduction or conclusion text.
The confidence score should be a number between 0 and 100, representing the confidence percentage.

[\{\{"edge": [source\_id, target\_id], "confidence": 85\}\},\{\{"edge": [source\_id, target\_id], "confidence": 80\}\}...]

\tcblower

\textbf{\textit{User Prompt}:}

Here is the data for the current analysis session.

[Agent Output Logs]

\{nodes\_block\}

Based on the criteria defined in the system instructions, please output the Top $k$ Highest Confidence Edges in raw JSON format, and sort them by Confidence in descending order.
Remember: The confidence score should be a number between 0 and 100, representing the confidence percentage.

[\{\{"edge": [source\_id, target\_id], "confidence": 85\}\},\{\{"edge": [source\_id, target\_id], "confidence": 80\}\}...]
\end{tcolorbox}
\caption{Prompt template for LLM-guided weak supervision.}
\label{tab:prompt_ws}
\end{table*}

\section{Prompts for Baselines}
\label{app:prompts Baselines}

\Cref{tab:prompt_bs} presents the exact prompts used to instruct baseline models to infer the communication topology of MAS.

\begin{table*}[t]
\centering
\begin{tcolorbox}[
    enhanced,
    width=\textwidth,
    boxrule=0.6pt,
    arc=2mm,
    left=4pt,right=4pt,top=4pt,bottom=4pt,
    title=\textbf{Prompt for LLM Inferring the Communication Topology of MAS},
    coltitle=black,
    colback=teal!1!white,
    colframe=teal!50!white,
    boxed title style={
      colback=teal!12!white,
      colframe=teal!28!black,
      boxrule=0.5pt,
      arc=2mm
    },
]
\textbf{\textit{System Prompt}:}

\textbf{Role}: You are an expert in Multi-Agent System topology analysis. 
Your core capability is Topology Inference: infer the hidden communication topology by analyzing textual interaction logs.

\textbf{Task}: Analyze the provided agent reasoning outputs to identify communication between agents. You must quantify the likelihood of a directed edge existing from Agent $i$ to Agent $j$ (Edge $[i \to j]$) using a Confidence Score (0-100\%).

Confidence Scoring Framework: Assign a score based on the strength of the evidence found in Agent $j$'s response relative to Agent $i$'s output:

\textbf{High Confidence (80-100\%): Explicit Reference \& Strict Execution.}

Explicit Citation: Agent $j$ explicitly mentions Agent $i$ (e.g., "Using the code provided above...", "Following the previously suggested approach...").

Hard Dependency: Agent $j$ executes code, runs a SQL query, or parses a specific data topology that was strictly defined or generated by Agent $i$. Agent $j$ cannot function without this specific asset.

\textbf{Medium Confidence (50-79\%): Content Dependency \& Semantic Alignment.}

Unique Information Flow: Agent $j$ solves a problem using unique information or a specific plan provided only by Agent $i$, even without explicit citation.

High Semantic Similarity: Agent $j$'s content exhibits high semantic overlap with Agent $i$. This includes reusing specific terminologies, variable names, or continuing a niche topic introduced by $i$, indicating a direct continuation of the context.

Logical Mapping: There is a clear "Question $\to$ Answer" or "Task $\to$ Solution" mapping between $i$ and $j$.

\textbf{Low Confidence (<50\%): Weak Logical Sequence.}

Agent $j$ and Agent $i$ have a weak logical sequence, the content is generic, or the solution could have been derived independently without Agent $i$'s specific input.

\textbf{Important}:

1. The MAS communication topology is directed, and edge direction must follow the temporal order in the provided textual interaction logs. 
Specifically, if Agent i's output appears earlier than Agent j's output in the logs, you may predict an edge [i -> j] but you must NOT predict [j -> i]. 
In other words, information can only flow forward in the log order (earlier agent -> later agent).

2. ID Consistency: Use the exact Agent IDs (e.g., 0, 1, 2) provided in the input.

3. Completeness: Output the confidence scores for all possible agent pairs; do not omit any edges, even those with low confidence.

4. Confidence Sorting: Output the edges sorted by Confidence in descending order, starting with the highest confidence edge.

5. No Markdown: Do not use markdown code blocks.

Provide a raw JSON list in the following format. Do not include any introduction or conclusion text.
The confidence score should be a number between 0 and 100, representing the confidence percentage.

[\{\{"edge": [source\_id, target\_id], "confidence": 85\}\},\{\{"edge": [source\_id, target\_id], "confidence": 80\}\}...]

\tcblower

\textbf{\textit{User Prompt}:}

Here is the data for the current analysis session.

[Agent Output Logs]

\{nodes\_block\}

Based on the criteria defined in the system instructions, please output the Top $k$ Highest Confidence Edges in raw JSON format, and sort them by Confidence in descending order.
Remember: The confidence score should be a number between 0 and 100, representing the confidence percentage.

[\{\{"edge": [source\_id, target\_id], "confidence": 85\}\},\{\{"edge": [source\_id, target\_id], "confidence": 80\}\}...]
\end{tcolorbox}
\caption{Prompt template for baseline models to infer the communication topology of MAS.}
\label{tab:prompt_bs}
\end{table*}

\end{document}